\documentclass[10pt,twocolumn,letterpaper]{article}

\usepackage{cvpr}              
\newcommand{\sysname}{DiT-BlockSkip}

\definecolor{cvprblue}{rgb}{0.21,0.49,0.74}
\usepackage[pagebackref,breaklinks,colorlinks,allcolors=cvprblue]{hyperref}
\usepackage{algorithm}
\usepackage{algorithmic}
\usepackage{multirow}
\usepackage{makecell}
\usepackage{graphicx}

\def\paperID{} 
\def\confName{CVPR}
\def\confYear{2026}

\title{Memory-Efficient Fine-Tuning Diffusion Transformers \\ via Dynamic Patch Sampling and Block Skipping}

\newcommand*{\affmark}[1][*]{\textsuperscript{#1}}
\newcommand\blfootnote[1]{%
  \begingroup
  \renewcommand\thefootnote{}\footnote{#1}%
  \addtocounter{footnote}{-1}%
  \endgroup
}

\author{
Sunghyun Park\affmark[1]$^{\ast}$\quad Jeongho Kim\affmark[1]$^{\ast}$\quad Hyoungwoo Park\affmark[1]\quad Debasmit Das\affmark[1]\\ 
Sungrack Yun\affmark[1]\quad Munawar Hayat\affmark[1]\quad Jaegul Choo\affmark[2]\quad Fatih Porikli\affmark[1]\quad Seokeon Choi\affmark[1]$^\dagger$\\
\affmark[1]Qualcomm AI Research$^\ddagger$ \; \affmark[2]KAIST\\
\texttt{\footnotesize\{sunpar, jeonghok, hwoopark, debadas, sungrack, fporikli, seokchoi\}@qti.qualcomm.com}\\
\texttt{\footnotesize jchoo@kaist.ac.kr}
\vspace{-0.3cm}
}

\begin{document}
\maketitle

\blfootnote{\hspace{-0.4cm}$^*$These two authors contributed equally to this work.}
\blfootnote{\hspace{-0.4cm}$^\dagger$Corresponding author.} 
\blfootnote{\hspace{-0.4cm}$^\ddagger$Qualcomm AI Research is an initiative of Qualcomm Technologies, Inc.\vspace{-1.3cm}}

\begin{abstract}
  Diffusion Transformers (DiTs) have significantly enhanced text-to-image (T2I) generation quality, enabling high-quality personalized content creation. 
  However, fine-tuning these models requires substantial computational complexity and memory, limiting practical deployment under resource constraints.
  To tackle these challenges, we propose a memory-efficient fine-tuning framework called DiT-BlockSkip, integrating timestep-aware dynamic patch sampling and block skipping by precomputing residual features.
  Our dynamic patch sampling strategy adjusts patch sizes based on the diffusion timestep, then resizes the cropped patches to a fixed lower resolution. 
  This approach reduces forward \& backward memory usage while allowing the model to capture global structures at higher timesteps and fine-grained details at lower timesteps.
  The block skipping mechanism selectively fine-tunes essential transformer blocks and precomputes residual features for the skipped blocks, significantly reducing training memory.
  To identify vital blocks for personalization, we introduce a block selection strategy based on cross-attention masking.
  Evaluations demonstrate that our approach achieves competitive personalization performance qualitatively and quantitatively, while reducing memory usage substantially, moving toward on-device feasibility (\textit{e.g.}, smartphones, IoT devices) for large-scale diffusion transformers.
\end{abstract}
\section{Introduction}

With the advent of large-scale text-to-image (T2I) generation frameworks, such as Stable Diffusion~\cite{ldm}, personalized content creation has surged, driving applications such as avatar generation and customized imagery~\cite{textual_inversion,dreambooth,customdiffusion,hyperdreambooth,elite,instantbooth}.
A common approach to personalization involves fine-tuning pre-trained T2I models using a few user-provided reference images~\cite{dreambooth,textual_inversion,customdiffusion}.
However, this fine-tuning process incurs substantial computational and memory overhead, especially large-scale models, limiting practical deployment under resource constraints.


\begin{figure*}
    \centering
    \includegraphics[width=0.9\linewidth]{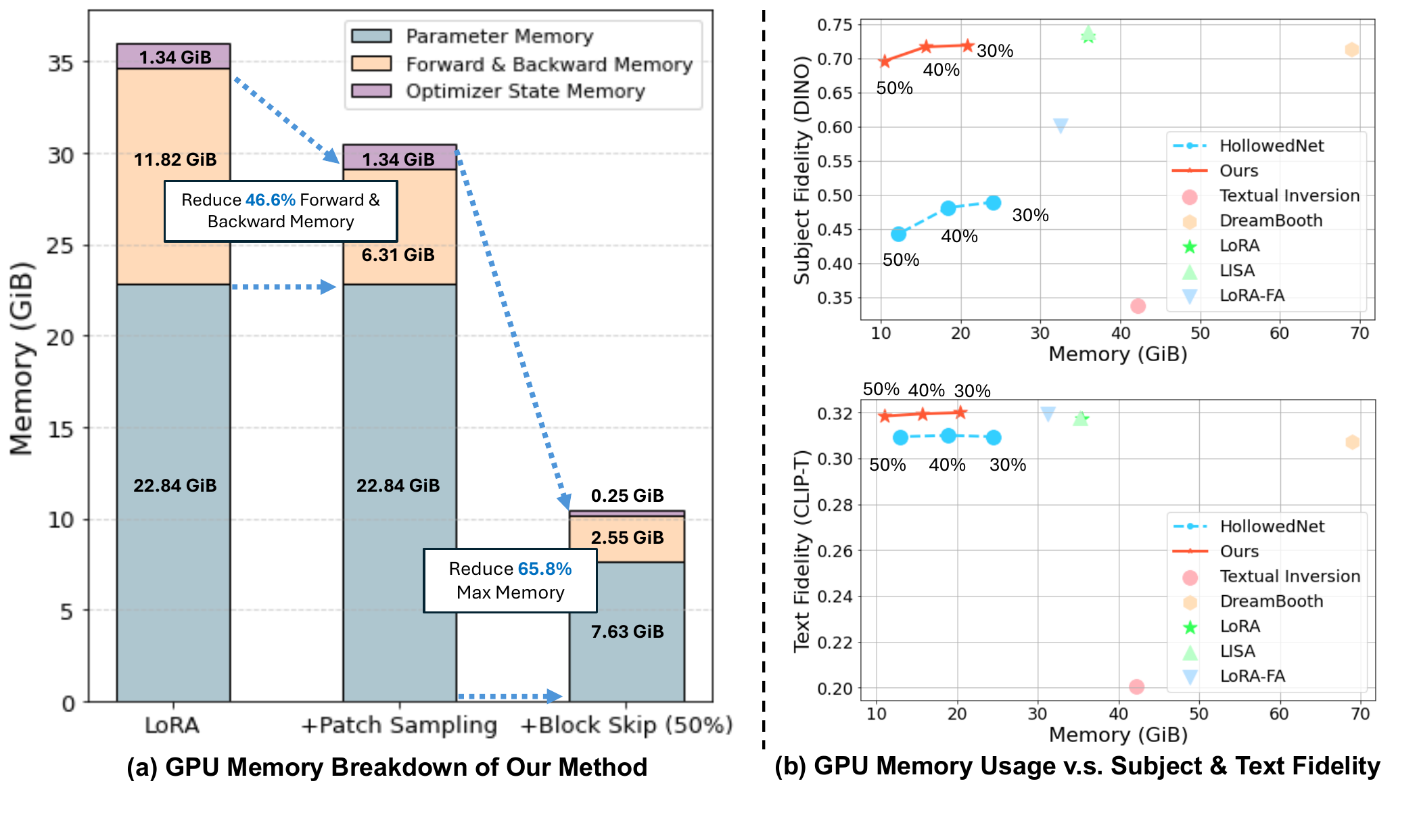}
    \vspace{-0.7cm}
    \caption{Memory consumption analysis during fine-tuning DiT (\textit{i.e.}, FLUX). (a) We propose~\sysname~including patch sampling and block skipping, which reducing the training memory usage significantly. (b) Comparison with baselines and~\sysname~in aspects of GPU memory usage and subject \& text fidelity.}
    \vspace{-0.5cm}
    \label{fig:teaser}
\end{figure*}

To mitigate these issues, parameter-Efficient Fine-Tuning (PEFT) methods, including LoRA~\cite{lora} and its variants~\cite{dora,vera,shira,lorafa}, freeze the original base model weights and only update small amounts of parameters.
However, these methods still require full backpropagation through the entire network, resulting in substantial memory overhead due to the excessive number of model parameters and activations involved. 
This remains a challenge even with quantization~\cite{qlora,qalora,peqa,tuneqdm}, which can further compromise precision.
Recently, several approaches have adopted gradient-free or zeroth-order optimization strategies~\cite{gradientfreeti,zoodip} to fine-tune models without relying on backpropagation, thereby significantly reducing memory consumption during fine-tuning.
However, these approaches often suffer from instability during optimization and require a large number of training iterations to converge~\cite{zoodip}.

With the emergence of diffusion transformer (DiT)~\cite{dit}-based T2I models~\cite{sd3,flux,pixart-sigma,sana}, recent architectures have dramatically increased in depth and capacity, enabling the generation of high-resolution images.
However, this advancement comes at the cost of significantly higher memory consumption during fine-tuning, posing challenges for practical deployment, particularly in resource-constrained scenarios.
Furthermore, the existing efficient fine-tuning techniques~\cite{hollowednet,gradientfreeti,tuneqdm,zoodip} have validated on U-Net-based T2I models, leaving DiT-based architectures relatively underexplored.
Therefore, achieving effective personalization with DiT-based T2I models necessitates further reductions in memory consumption.

%

To address these challenges, we propose a novel scheme, called \textbf{\sysname}, that significantly reduces the training memory overhead while preserving output quality (Fig.~\ref{fig:teaser}).
Our method consists of two key components: (1) dynamic patch sampling and (2) block skipping with residual feature precomputation.
The first component reduces forward \& backward memory significantly by using on low-resolution patch images, wherein the patch size is dynamically adjusted based on the diffusion timestep.
Here, we first crop the high-resolution images based on the adjusted patch sizes, and then we resize these patches of varying sizes into a fixed low-resolution size (Sec.~\ref{sec:patch_sampling}). 
The core intuition is that higher timesteps are associated with learning global structure, while lower timesteps are more effective for capturing fine-grained details~\cite{diffusion_curriculum,progressive}.
By aligning crop ratios with these characteristics, the model can effectively learn both structural and detailed information even from low-resolution inputs, mimicking the representational benefits of high-resolution training.
The second component, block skipping with residual feature precomputation, selectively updates only a subset of essential transformer blocks for personalization during fine-tuning (Sec.~\ref{sec:block_skipping}). 
For skipping blocks, precomputed residual feature maps are reused, allowing the model to bypass unnecessary forward and backward computations. 
This strategy is motivated by the observation that full fine-tuning of DiT is often redundant, and learning can be preserved by focusing on a strategically chosen subset of critical blocks. 
Moreover, we introduce a block selection strategy based on cross-attention masking to identify critical blocks for personalization.
Together, the proposed methods significantly reduce the training memory usage while maintaining the personalization performance.
Our contributions are summarized as follows:
\begin{itemize}
    \setlength{\itemsep}{0pt}
    \setlength{\parskip}{2pt}
    \setlength{\parsep}{2pt}
    \item To reduce the training memory, we introduce a patch sampling strategy that dynamically adjusts the patch sizes based on diffusion timesteps, learning both global structures and fine-grained details from low-resolution images.
    \item We propose a method that selectively fine-tunes essential transformer blocks for personalization while utilizing pre-extracted residual features for the skipped blocks. This technique maintains personalization quality while minimizing computational overhead.
    \item The experiments demonstrate that our method produces competitive performance while significantly reducing training memory usage. 
\end{itemize}

\section{Related Work}


\begin{figure*}[t!]
    \centering
    \includegraphics[width=0.9\linewidth]{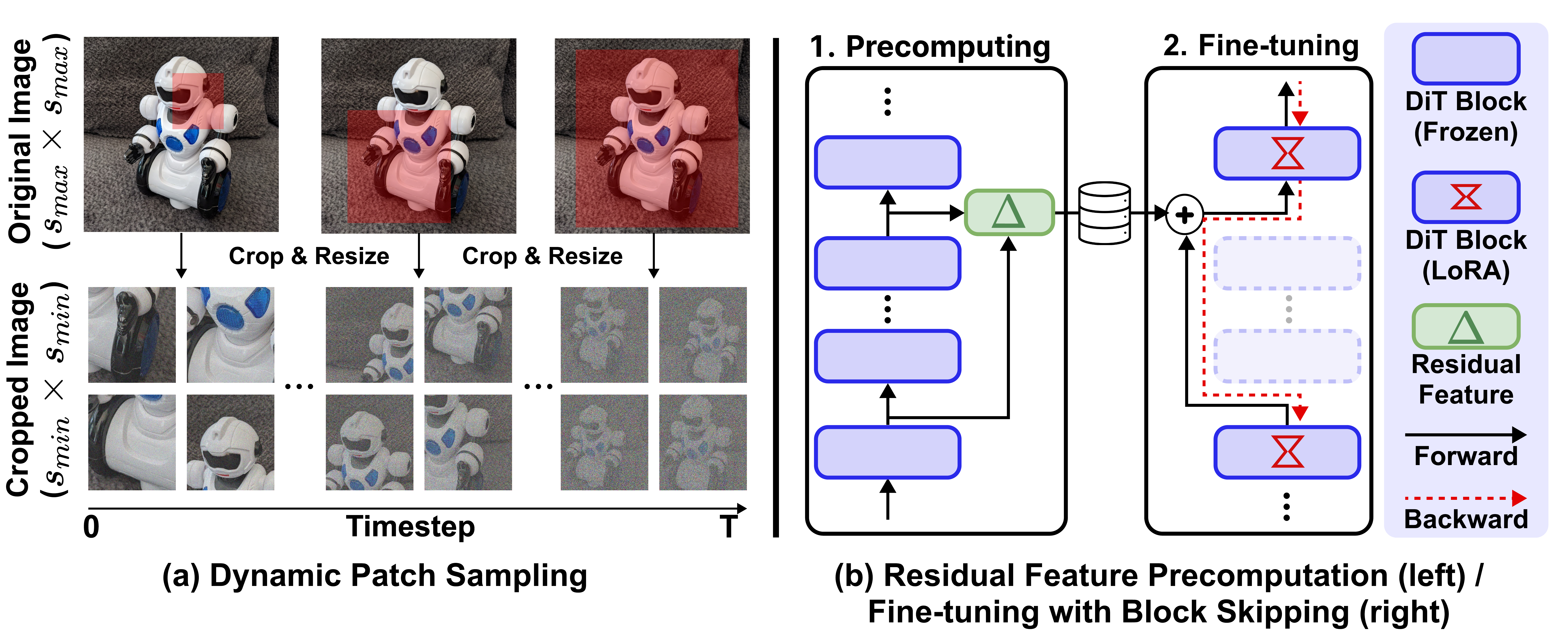}
    \vspace{-0.2cm}
    \caption{Overview of the proposed method. (a) The dynamic patch sampling applies different patch sizes for each diffusion timestep, enabling the model to learn both global structure and fine-grained details depending on the noise level. Cropped patches of various sizes are resized to the same fixed resolution ($s_{min} \times s_{min}$) before being fed into the model. (b) Block skipping with residual features for DiT stores the $\Delta$ value between the input and output of the consecutive skipped blocks in the precomputation stage. During fine-tuning, these residual feature is added to the input of the skipped blocks, while the remaining blocks are updated end-to-end.}
    \vspace{-0.5cm}
    \label{fig:main}
\end{figure*}

\noindent\textbf{Personalization of Diffusion Models.}
Personalization in diffusion models aims to adapt pre-trained models to synthesize images of user-defined concepts using reference images. 
Training-based methods~\cite{dreambooth,textual_inversion,customdiffusion,classdiffusion,hyperdreambooth,pg} typically fine-tune the parameters or embeddings to learn the target concepts. 
While effective, these methods often require substantial computational resources and memory during fine-tuning, limiting their applicability on edge devices.
In contrast, encoder-based approaches~\cite{instantbooth,elite,hyperdreambooth} leverage large-scale datasets to pre-train additional encoders that enable personalization via a single forward pass during inference.
However, they require the encoder to be separately trained for each pre-trained T2I model and introduce additional encoder parameters, limiting their flexibility and scalability.
Training-free approaches~\cite{jedi,freecustom,consistory} inject personalization cues during inference through carefully designed operations, enabling faster personalization without training.
Nonetheless, these techniques often increase inference-time latency and memory usage, which can be prohibitive in resource-constrained environments. 
In this paper, we propose an efficient fine-tuning strategy that minimizes memory consumption during training, enabling effective personalization without incurring additional memory or latency overhead at inference time.

\noindent\textbf{Memory-Efficient Fine-Tuning for Diffusion Models.}
Enabling fine-tuning large diffusion models on resource-constrained devices necessitates memory-efficient training strategies.
Parameter-efficient fine-tuning (PEFT) methods~\cite{lora,dora,shira,vera,para,lorafa} reduce the number of trainable parameters, thereby decreasing optimizer state memory usage during training.
However, most PEFT methods still rely on full backpropagation, which can be memory-intensive due to the base model parameter and activation memory.
Recent advancements~\cite{tuneqdm,zoodip,hollowednet,lisa,mixopt} have explored alternative approaches to mitigate memory consumption during fine-tuning. 
For instance, ZOODiP~\cite{zoodip} advances the quantized diffusion models by employing zeroth-order optimization within textual inversion, thereby eliminating the need for backpropagation entirely.
However, ZOODiP often suffers from optimization instability, making them require an excessive number of training steps to converge (\textit{e.g.}, 30,000 steps).
HollowedNet~\cite{hollowednet} exploits architectural priors in U-Net backbones by fine-tuning only a subset of layers. 
However, this method is inherently specialized for U-Net architectures and does not generalize well to DiT-based models, which are becoming increasingly prevalent. 
To overcome these limitations, we propose~\sysname, enabling effective personalization with significantly reduced memory.

\noindent\textbf{Layer Skipping.}
To reduce the computational cost of inference, previous research has investigated skipping layers dynamically during inference~\cite{blockdrop,layerskip}.
More recently, diffusion models, requiring multiple timesteps for image generation, have adopted inference-time acceleration methods~\cite{deepcache,delta,learning-to-cache,faster,cacheme,token,videocache,reusing} by skipping layers and reusing the features.
However, these approaches are inherently designed for inference and cannot be directly applied to the training involving parameter updates.
The layer skipping during training remains underexplored, particularly in scenarios involving personalization.
Motivated by the potential of reducing memory and computational overhead, we propose a fine-tuning strategy for DiT that selectively skips transformer blocks during training.
Our method identifies layers most critical for personalization and skips non-essential blocks, reducing memory usage while preventing performance degradation.


\section{Method}

\begin{figure*}[t!]
    \centering
    \includegraphics[width=0.9\linewidth]{figures/skip_selection.pdf}
    \vspace{-0.25cm}
    \caption{Qualitative results and mean semantic distances are reported for LoRA fine-tuned FLUX on the CustomConcept101 dataset~\cite{customdiffusion}, evaluated across 30 randomly selected classes. We mask the attention scores from image (query) to text (key) within the joint attention across 14 consecutive blocks. Masking mid-level blocks causes a significant drop in similarity compared to full attention.}
    \vspace{-0.4cm}
    \label{fig:skip_selection}
\end{figure*}

\subsection{Dynamic Patch Sampling}
\label{sec:patch_sampling}

Diffusion models typically require high-resolution images for fine-tuning to generate high-resolution images and capture fine-grained details. 
However, fine-tuning at high resolutions significantly increases memory consumption. 
To address this issue, we propose a dynamic patch sampling strategy that effectively balances memory efficiency with personalization quality.
Specifically, instead of processing the full-resolution input images throughout the fine-tuning process, we resize images to the fixed lower resolutions (\textit{e.g.}, $256 \times 256$) using a dynamic patch sampling. 

As shown in Fig.~\ref{fig:main} (a), this strategy adjusts the patch sizes based on the current diffusion timestep, enabling the model to focus on different aspects of the target concept as training progresses.
At higher timesteps, where the input is heavily perturbed with noise, a larger patch is sampled to emphasize learning the global structure of the subject. 
Conversely, at lower timesteps, where noise is reduced, a smaller patch is sampled to capture fine-grained local details. 
Formally, given a sampled timestep $t$, we define a patch size function $f(s_\text{min}, s_\text{max}, t)$ that determines the portion of the image to retain:
\begin{equation}
    f(s_\text{min}, s_\text{max}, t) = s_{\text{min}} + \frac{t}{T} \cdot (s_{\text{max}} - s_{\text{min}}), 0 \leq t \leq T,
\end{equation}
where $s_{\text{min}}$ and $s_{\text{max}}$ denote the minimum and maximum patch size, respectively, and $T$ is the maximum diffusion timestep. 
This decay ensures that the cropping area gradually shrinks as the model transitions from coarse structure learning to fine detail refinement. 
For the implementation details, the patch size does not vary in a fully continuous linear manner. 
Instead, the final patch size is discretized based on the resolution downscaling factor of the VAE encoder (\textit{e.g.}, 16), and is determined by rounding to the nearest multiple of the encoder’s resolution scale.


After cropping, the sampled image patch is resized to the same resolution as $s_{\text{min}} \times s_{\text{min}}$ and used for training. 
Although the training resolutions are identical, the patches may capture either global structures or fine-grained details of the image depending on the timestep, allowing the model to learn both global and local features through dynamic patch sampling.
This method significantly reduces the forward \& backward memory without modifying the model architecture or the denoising objective, making it straightforward to integrate into diffusion model fine-tuning. 

\subsection{Block Skipping with Residual Feature}
\label{sec:block_skipping}

We propose a block skipping strategy for fine-tuning DiTs that reduces model parameters while preserving personalization capacity. 
As illustrated in Fig.~\ref{fig:main} (b), our method selectively skips a sequence of consecutive transformer blocks. 
Specifically, when skipping $l$ consecutive blocks starting from the $i$-th block, we directly replace the $(i+l)$-th feature $f_{i+l}$ with $f_i$.
However, as shown in Fig.~\ref{fig:ablation_crop_and_skip_position} (b), naive skipping introduces a mismatch between the forward paths during training and inference, leading to significant degradation in personalization performance.
To mitigate this issue, we introduce a training framework consisting of: (1) selecting skipped blocks, (2) precomputing residual feature, and (3) fine-tuning selected blocks.

\noindent\textbf{Selection Strategy for Block Skipping.}
T2I diffusion models based on the U-Net architecture~\cite{ldm,sdxl,dalle2} exhibit a well-known hierarchical pattern due to progressive spatial downsampling: shallow blocks capture low-level features, while deeper ones learn high-level semantics such as object shapes.
In contrast, DiTs~\cite{dit,flux,pixart-sigma,sd3} maintain a constant spatial resolution, making the functional roles of individual blocks less interpretable, especially for personalization.

Here, we aim to explore an important and intriguing question: \emph{How can we identify the blocks that are crucial for personalization?}
To explore this, we fine-tuned a DiT model using LoRAs~\cite{lora} modules inserted into all blocks.
During inference, we applied cross-attention masking from image (query) to text (key) across 14 consecutive transformer blocks with a stride of 4, effectively suppressing prompt influence in distinct block segments while avoiding redundant computation due to excessive overlap. 

Interestingly, as shown in Fig.~\ref{fig:skip_selection}(a), masking early or late blocks yields results nearly identical to the full model (\textit{i.e.}, without masking).
However, masking the mid-level blocks causes the subject to disappear, indicating mid-level blocks play a critical role in encoding subject-specific information.

To quantify this, we compute semantic distances between image embeddings from images generated by the full model and those with masked blocks.
The image embeddings are extracted using a pre-trained encoder (\textit{i.e.}, DINO~\cite{dino}), and semantic distance as: $D(x, y) = 1 - \text{sim}(x, y)$, where $\text{sim}(x, y)$ denotes the similarity between embeddings extracted from images $x$ and $y$.

Fig.~\ref{fig:skip_selection}(b) shows a quantitative analysis of semantic distances between images generated with full attention and those with cross-attention masking applied to block segments, evaluated on 30 randomly selected classes from the CustomConcept101 dataset~\cite{customdiffusion}. 
The results consistently show that masking mid-level blocks leads to the largest semantic difference, further validating their importance for subject representation.

Motivated by these observations, we propose a strategy to keep mid-level blocks in the DiT model. 
Specifically, we skip the first $n$ blocks and the last $m$ blocks from the end of the DiT model. 
For each LoRA fine-tuned DiT model personalized to a subject, we generate:
\begin{itemize}
    \setlength{\itemsep}{0pt}
    \setlength{\parskip}{2pt}
    \setlength{\parsep}{2pt}
    \item $\hat{x}_n$: image with first $n$ cross-attention blocks masked
    \item $\tilde{x}_m$: image with last $m$ cross-attention blocks masked
\end{itemize}
Given a reference image $x_g$ generated without cross-attention masking, our goal is to find a pair $(n^*, m^*)$ that minimizes the influence of the text prompt, as measured by the semantic distance to $x_g$. 
\begin{equation}
    (n^*, m^*) = \operatorname*{arg\,min}_{n+m=k} \sum_{j=1}^{N} \left(D(x_g^{(j)}, \hat{x}_n^{(j)}) + D(x_g^{(j)}, \tilde{x}_m^{(j)})\right).
\end{equation}
Here, $N$ is the number of DiT-based models fine-tuned with LoRA, each trained on a different subject (\textit{e.g.,} 30 instances of CustomConcept101) and denoted as $\mathcal{M}_j$ ($j \in [1, N]$). All images, $x_g^{(j)}$, $\hat{x}_n^{(j)}$, and $\tilde{x}_m^{(j)}$, are generated from the $j$-th DiT model ($\mathcal{M}_j$).
The constraint $n + m = k$ controls the total number of skipped blocks. The entire procedure for selecting skip indices is summarized in Algorithm~\ref{alg:layer_selection}.

A key advantage of this approach is its computational efficiency.
Once distances $D(x_g, \hat{x}_n)$ and $D(x_g, \tilde{x}_m)$ are precomputed for all $n, m \in {1, \dots, L-1}$, where $L$ denotes the total number of transformer blocks in the DiT model, the skip pair $(n^*, m^*)$ can be quickly retrieved for any given total number of skip blocks $k$.

\noindent\textbf{Precomputing Residual Feature.} 
After selecting the skipped blocks, we compute and store the residual features defined as the difference between the input to the first skipped block $f_i$ and the output $f_{i+l}$ of the last skipped block.
This residual feature $\Delta f_{i,i+l}=f_{i+l} - f_i$ (green box in Fig.~\ref{fig:main} (b)) serves to preserve the information of the skipped blocks, enabling backpropagation even with the modified network structure.
The skipped blocks correspond to the first $n^*$ and the last $m^*$ blocks, for which we precompute the features $\Delta f_{0, n^*}$ and $\Delta f_{L-m^*, L}$, respectively. 
Residual features are extracted in the same number as the training iterations.
Considering that image generation during inference typically involves 20 to 50 forward passes through the DiT model, the computational overhead introduced by feature precomputing remains relatively modest.

\begin{algorithm}[t]
\caption{Select Skip Indices via Cross-Attention Mask}
\label{alg:layer_selection}
\begin{algorithmic}[1]
\REQUIRE Set of $N$ LoRA fine-tuned DiT models $\{\mathcal{M}_1, \dots, \mathcal{M}_N\}$, prompt $p$, total skip count $k$
\STATE Initialize empty candidate list
\FOR{$n = 1$ to $k - 1$}
    \STATE $m \leftarrow k - n$, $D \leftarrow 0$
    \FOR{$j = 1$ to $N$}
        \STATE $x^{(j)}_g \leftarrow$ image from $\mathcal{M}_i$ with full attention
        \STATE $\hat{x}^{(j)}_n \leftarrow$ image from $\mathcal{M}_j$ with cross-attention masked in first $n$ blocks
        \STATE $\tilde{x}^{(j)}_m \leftarrow$ image from $\mathcal{M}_j$ with cross-attention masked in last $m$ blocks
        \STATE $D \leftarrow D + D(x^{(j)}_g, \hat{x}^{(j)}_n) + D(x^{(j)}_g, \tilde{x}^{(j)}_m)$
    \ENDFOR
    \STATE Append $(n, m, D)$ to candidate list
\ENDFOR
\STATE $(n^*, m^*) \leftarrow \arg\min_{n,m} D$ from candidate list
\RETURN $(n^*, m^*)$
\end{algorithmic}
\end{algorithm}

\noindent\textbf{Fine-tuning Unskipped Blocks.}
LoRA~\cite{lora} is injected to the unskipped blocks. 
To fine-tune the LoRA weights, we employ the conditional flow matching loss~\cite{sd3,lipman2022flow}.
The previously precomputed residual feature is then loaded from storage and added to the updated input $f^\prime_i$ due to the updated LoRA weights, forming the input to $(i+l)$-th block as:
\begin{equation}
    f^\prime_{i+l} = f^\prime_{i} + \Delta{f_{i,i+l}} = f^\prime_{i} + (f_{i+l} - f_i).
\end{equation}
In the fine-tuning process, the selected skipped block parameters are offloaded from the GPU. 
In their place, precomputed residual features $\Delta f_{i, i+l}$ are used, which significantly reduces both parameter and forward \& backward memory usages, as shown in Fig.~\ref{fig:teaser}. 
Moreover, since only the LoRAs applied to specific blocks are updated, the optimizer state memory is also substantially reduced.

\begin{table*}[t!]
\centering
\scriptsize 
\caption{Comparison with baselines based on FLUX and SANA. Ratio refers to skip ratio. Resolution denotes the training resolution. Inference resolution is fixed at 1024$\times$1024.}
\vspace{-0.3cm}
\resizebox{0.9\textwidth}{!}{
\begin{tabular}{l|cc|ccc|cc|ccc}
    \toprule
    \multirow{2}{*}{\textbf{Method}} & \multicolumn{5}{c|}{\textbf{FLUX}} & \multicolumn{5}{c}{\textbf{SANA}} \\
    \cmidrule(lr){2-6} \cmidrule(lr){7-11}
    & Ratio & Resolution & DINO & CLIP-I & CLIP-T & Ratio & Resolution & DINO & CLIP-I & CLIP-T \\
    \midrule
    TI          & -- & 512$\times$512  & 0.3384 & 0.6140 & 0.2005 
                & -- & 1024$\times$1024 & 0.4185 & 0.6823 & 0.2334 \\
    DreamBooth  & -- & 512$\times$512  & 0.7131 & 0.8122 & 0.3073 
                & -- & 1024$\times$1024 & 0.6270 & 0.7612 & \textbf{0.3294} \\
    LoRA        & -- & 512$\times$512  & \underline{0.7324} & \underline{0.8146} & 0.3173 
                & -- & 1024$\times$1024 & \underline{0.7374} & \textbf{0.8108} & \underline{0.3254} \\
    LISA        & -- & 512$\times$512  & \textbf{0.7387} & \textbf{0.8194} & 0.3177 
                & -- & 1024$\times$1024 & 0.5226 & 0.7171 & 0.3188 \\
    LoRA-FA     & -- & 512$\times$512  & 0.6017 & 0.7433 & 0.3191 
                & -- & 1024$\times$1024 & 0.4692 & 0.6892 & 0.3224 \\
    \midrule
    Hollowed    & 30\% & 512$\times$512 & 0.4899 & 0.7031 & 0.3094 
                & 30\% & 1024$\times$1024 & 0.7208 & 0.7952 & 0.3149 \\
    Hollowed    & 40\% & 512$\times$512 & 0.4819 & 0.7112 & 0.3100 
                & 40\% & 1024$\times$1024 & 0.6794 & 0.7784 & 0.3067 \\
    Hollowed    & 50\% & 512$\times$512 & 0.4435 & 0.6930 & 0.3094 
                & 50\% & 1024$\times$1024 & 0.6301 & 0.7540 & 0.3064 \\
    \midrule
    Ours        & 30\% & 256$\times$256 & \underline{0.7194} & 0.8036 & \textbf{0.3199} 
                & 30\% & 512$\times$512 & \textbf{0.7388} & 0.8043 & 0.3240 \\
    Ours        & 40\% & 256$\times$256 & 0.7171 & 0.8034 & \underline{0.3194} 
                & 40\% & 512$\times$512 & 0.7333 & \underline{0.8086} & 0.3211 \\
    Ours        & 50\% & 256$\times$256 & 0.6963 & 0.7877 & 0.3184 
                & 50\% & 512$\times$512 & 0.7277 & 0.8034 & 0.3177 \\
    \bottomrule
\end{tabular}
}
\vspace{-0.3cm}
\label{table:qualitative}
\end{table*}

\begin{figure*}[t!]
    \centering
    \begin{subfigure}[t]{0.48\linewidth}
        \centering
        \includegraphics[width=\linewidth]{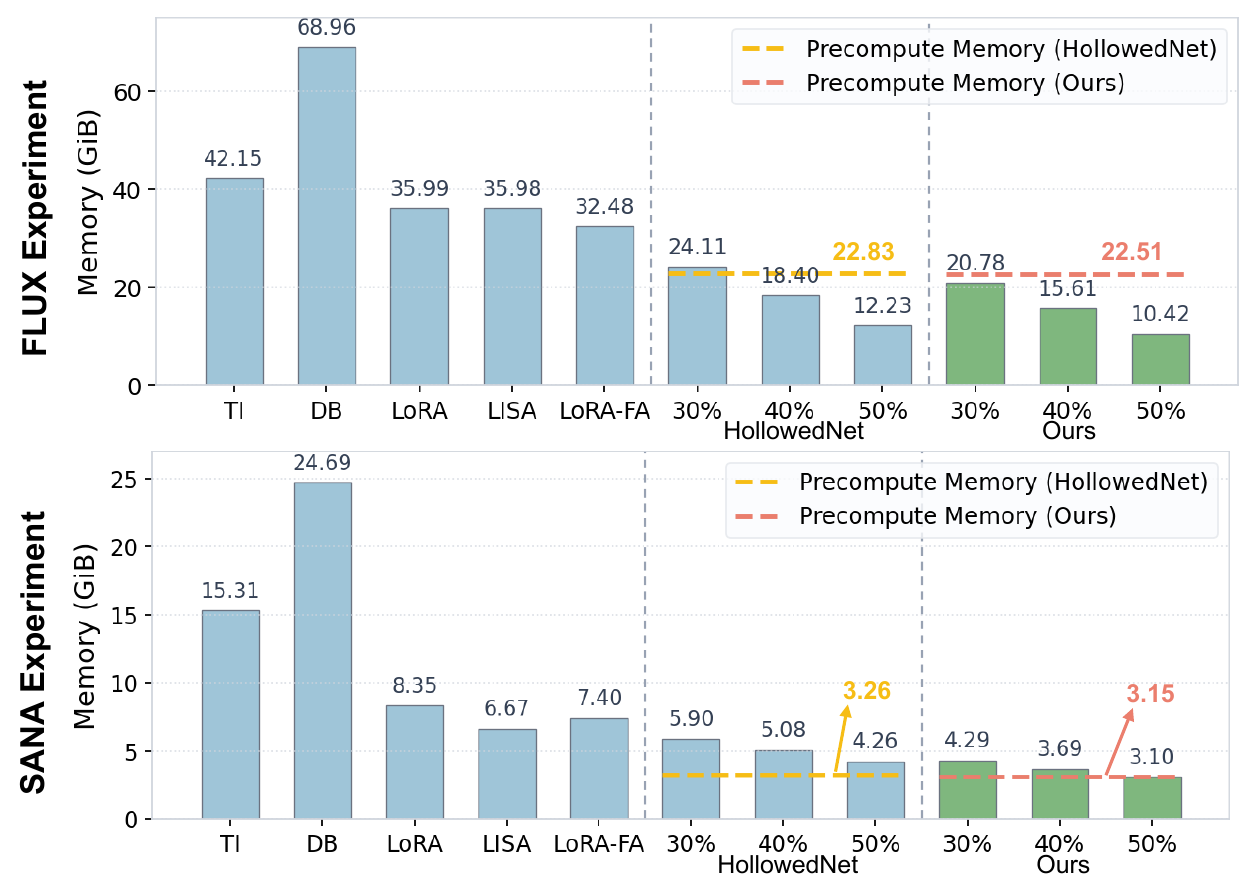}
        \caption{Training Memory}
        \label{fig:memory}
    \end{subfigure}
    \hfill
    \begin{subfigure}[t]{0.485\linewidth}
        \centering
        \includegraphics[width=\linewidth]{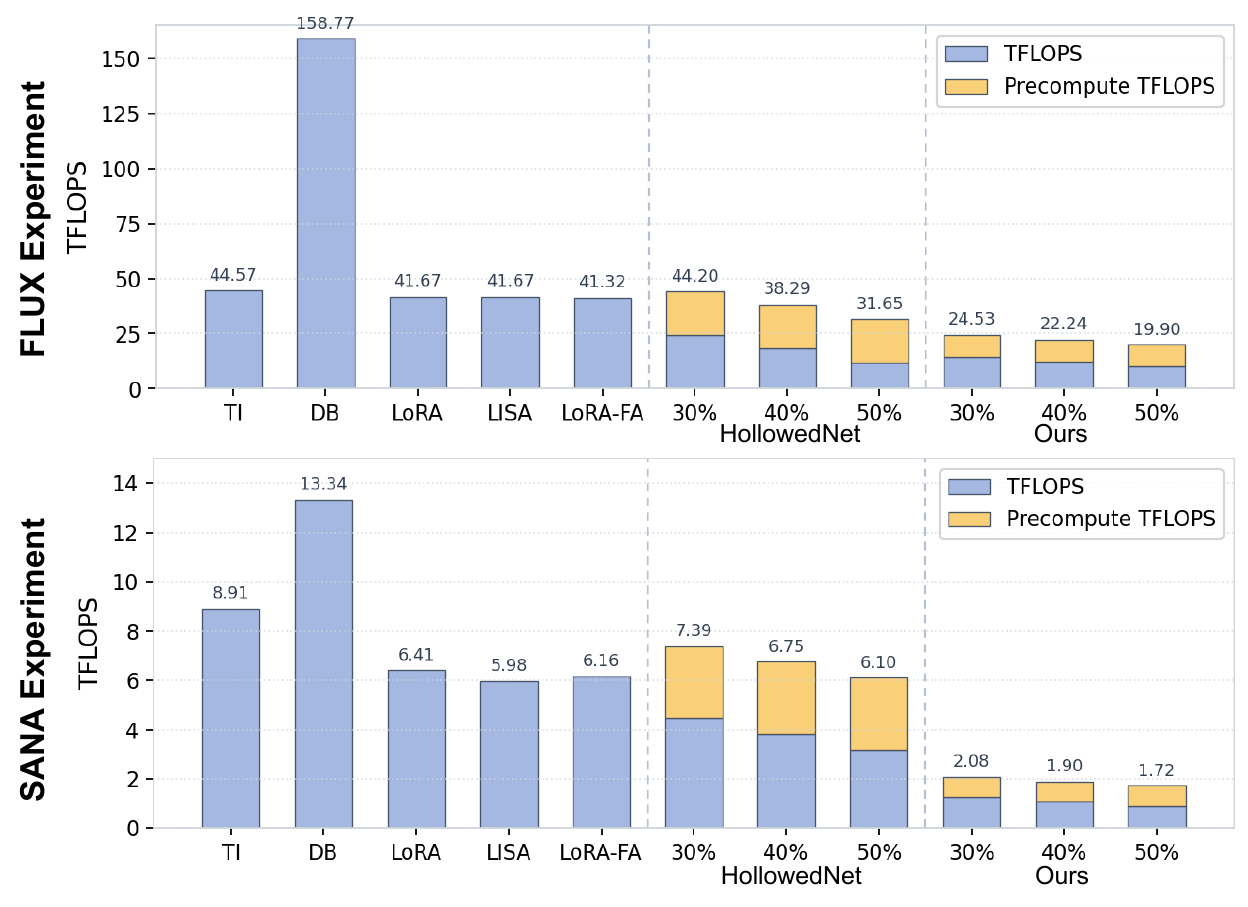}
        \caption{TFLOPS}
        \label{fig:tflops}
    \end{subfigure}
    \vspace{-0.4cm}
    \caption{Comparison of memory and TFLOPS.}
    \vspace{-0.4cm}
    \label{fig:memory_tflops}
\end{figure*}

\section{Experiments}\label{sec:experiment}
\noindent\textbf{Experimental Settings.}
We conducted experiments using FLUX.1-dev~\footnote{https://huggingface.co/black-forest-labs/FLUX.1-dev\vspace{-0.6cm}} and SANA~\cite{sana} as the base T2I models.
We conduct experiments using a total of 30 subjects in DreamBooth dataset~\cite{dreambooth}, where each subject contains 4-6 images.
Subjects are divided into living beings and objects, with 25 class-specific prompts assigned accordingly. 
For evaluation, four images are generated per subject per prompt using different random seeds.
For block skip with precomputed residual features, we experimented with skipping 30\%, 40\%, and 50\% of the DiT blocks.
Detailed information on the indices of the skipped blocks and experimental settings is provided in the supplementary.


\begin{figure*}[t!]
    \centering
    \includegraphics[width=0.95\linewidth]{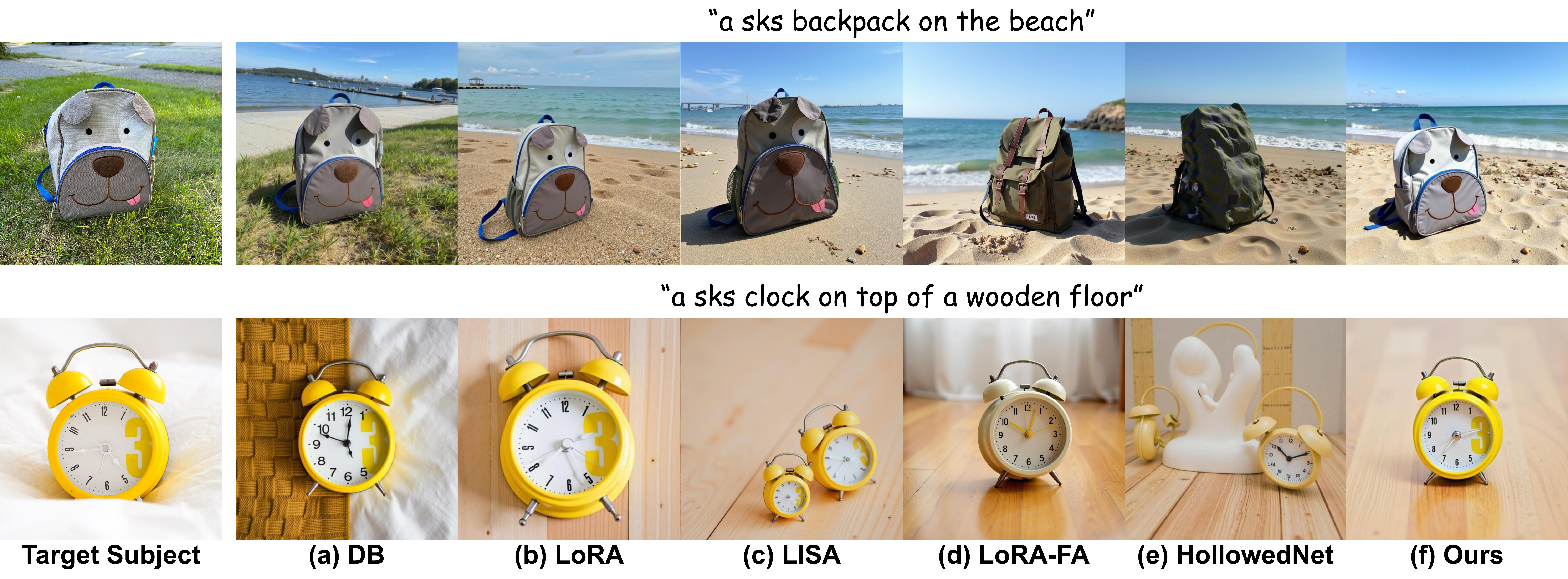}
    \vspace{-0.3cm}
    \caption{Qualitative comparison with baselines on the DreamBooth dataset.}
    \vspace{-0.3cm}
    \label{fig:qualitative}
\end{figure*}

\subsection{Results}

\noindent\textbf{Subject and Text Alignment.}
To evaluate personalization performance, we measure image-text alignment using three metrics: DINO~\cite{dino} and CLIP-I~\cite{clip} for subject fidelity, and CLIP-T~\cite{clip} for text prompt fidelity.
Our method integrates dynamic patch sampling and block skipping into LoRA, aiming to reduce training memory while preserving LoRA-level performance.
Table~\ref{table:qualitative} show quantitative comparisons across FLUX and SANA across various fine-tuning baselines, including personalization methods (Textual Inversion (TI)~\cite{textual_inversion}, DreamBooth (DB)~\cite{dreambooth}, LoRA~\cite{lora}) and memory-efficient fine-tuning approaches (LISA~\cite{lisa}, LoRA-FA~\cite{lorafa}, HollowedNet~\cite{hollowednet}).
Since HollowedNet was originally designed for U-Net architectures, we adapt it to DiTs by incorporating our block skipping with residual feature precomputation.
Our method consistently achieves performance comparable to LoRA, whereas LISA and LoRA-FA show degraded performance on SANA, and HollowedNet underperforms on FLUX. 
This performance drop is attributed to the fact that LISA and LoRA-FA are designed for LLMs and thus better suited for large-scale models like FLUX, but less effective for efficient models like SANA.
In contrast, HollowedNet applies empirically driven block skipping, which leads to excessive skipping of critical blocks and causes notable performance drops when naively applied to DiT.
The original HollowedNet paper also reported significant performance drops when skipping beyond a certain percentage of blocks, indicating difficulty in accurately identifying critical blocks.
Notably, TI~\cite{textual_inversion} performed poorly, even when applied to FLUX using the official Diffusers code, expecting the limited performance of gradient-free textual inversion~\cite{gradientfreeti} and ZOODIP~\cite{zoodip} to DiTs.

\begin{table}[t!]
\centering
\caption{Ablation study on FLUX.}
\vspace{-0.3cm}
\small
\resizebox{\linewidth}{!}{%
\setlength{\tabcolsep}{4pt}
\begin{tabular}{l|cc|ccc}
    \toprule
    \textbf{Method} & \textbf{Ratio} & \textbf{Resolution} & \textbf{DINO} & \textbf{CLIP-I} & \textbf{CLIP-T} \\
    \midrule
    \multicolumn{6}{l}{\textbf{Ablation Study on Dynamic Patch Sampling}} \\
    \midrule
    LoRA             & -- & 512$\times$512 & \textbf{0.7324} & \textbf{0.8146} & 0.3173 \\
    + Resize         & -- & 256$\times$256 & 0.7164 & 0.8044 & 0.3176 \\
    + Patch Sampling & -- & 256$\times$256 & 0.7253 & 0.8099 & \textbf{0.3196} \\
    \midrule
    \multicolumn{6}{l}{\textbf{Ablation Study on Residual Feature Precomputing}} \\
    \midrule
    \multirow{3}{*}{+ Block Skipping} & 30\% & 512$\times$512 & 0.4313 & 0.6815 & 0.3047 \\
                                      & 40\% & 512$\times$512 & 0.4338 & 0.6817 & 0.3051 \\
                                      & 50\% & 512$\times$512 & 0.4301 & 0.6794 & 0.3047 \\
    \midrule
    \multirow{3}{*}{\makecell{+ Block Skipping\\+ Residual Feature}}   & 30\% & 512$\times$512 & 0.7282 & \textbf{0.8148} & 0.3154 \\
                                      & 40\% & 512$\times$512 & \textbf{0.7303} & 0.8117 & 0.3145 \\
                                      & 50\% & 512$\times$512 & 0.7150 & 0.8035 & \textbf{0.3182} \\
    \midrule
    \multicolumn{6}{l}{\textbf{Ablation Study of Skipped Layer Positions}} \\
    \midrule
    First 50\% Blocks  & 50\% & 512$\times$512 & 0.6651 & 0.7646 & \textbf{0.3193} \\
    Last 50\% Blocks   & 50\% & 512$\times$512 & 0.4808 & 0.7111 & 0.3090 \\
    Our Strategy       & 50\% & 512$\times$512 & \textbf{0.7150} & \textbf{0.8035} & 0.3182 \\
    \bottomrule
\end{tabular}
}
\vspace{-0.5cm}
\label{table:ablation_flux}
\end{table}

\noindent\textbf{Training Memory and Computational Costs.}
Fig.~\ref{fig:memory} and~\ref{fig:tflops} show training memory measured with PyTorch in bfloat16 and TFLOPs for a single training iteration.
Specifically, when the text encoder and VAE are not involved in training, their outputs are precomputed and they can be offloaded to the CPU. 
Training memory is defined as the total memory consumed by the DiT, including parameters, optimizer states, and forward/backward memory.
FLUX and SANA with LoRA consume 22.84 GiB and 3.50 GiB of parameter memory, respectively.
Thus, the baselines including LoRA~\cite{lora}, LISA~\cite{lisa}, and LoRA-FA~\cite{lorafa} offer limited memory savings as the base model size remains unchanged.
In contrast, HollowedNet~\cite{hollowednet} and our method reduce memory effectively by offloading some blocks of the base model via block skipping. 
Compared to HollowedNet, dynamic patch sampling with low-resolution training is particularly effective in reducing TFLOPs. 
As shown in Fig.~\ref{fig:memory}, we also report the peak training memory when precomputing features. 
Notably, during training, our method can reduce memory usage even more than simply forwarding the model. 
Here, since features are precomputed without LoRA, precomputing memory is smaller than parameter memory that is measured after LoRA injection.
Furthermore, Fig.~\ref{fig:tflops} shows that our method achieves significantly lower TFLOPS compared to other approaches, even when precomputation is included.
Consequently, our approach demonstrates the most efficient training cost in terms of both memory usage and computational overhead.

\begin{table}[t!]
\centering
\small
\caption{User preference on subject fidelity and text fidelity.}
\vspace{-0.3cm}
\begin{tabular}{l|ccc}
    \toprule
    \textbf{Metric} & LoRA & Tie & Ours \\
    \midrule
    \textbf{Subject $\uparrow$} & 40.0\% & 23.3\% & 37.7\% \\
    \textbf{Text $\uparrow$}    & 29.4\% & 25.0\% & 45.6\% \\
    \bottomrule
\end{tabular}
\vspace{-0.5cm}
\label{table:userstudy}
\end{table}


\begin{figure*}[t!]
    \centering
    \includegraphics[width=0.92\linewidth]{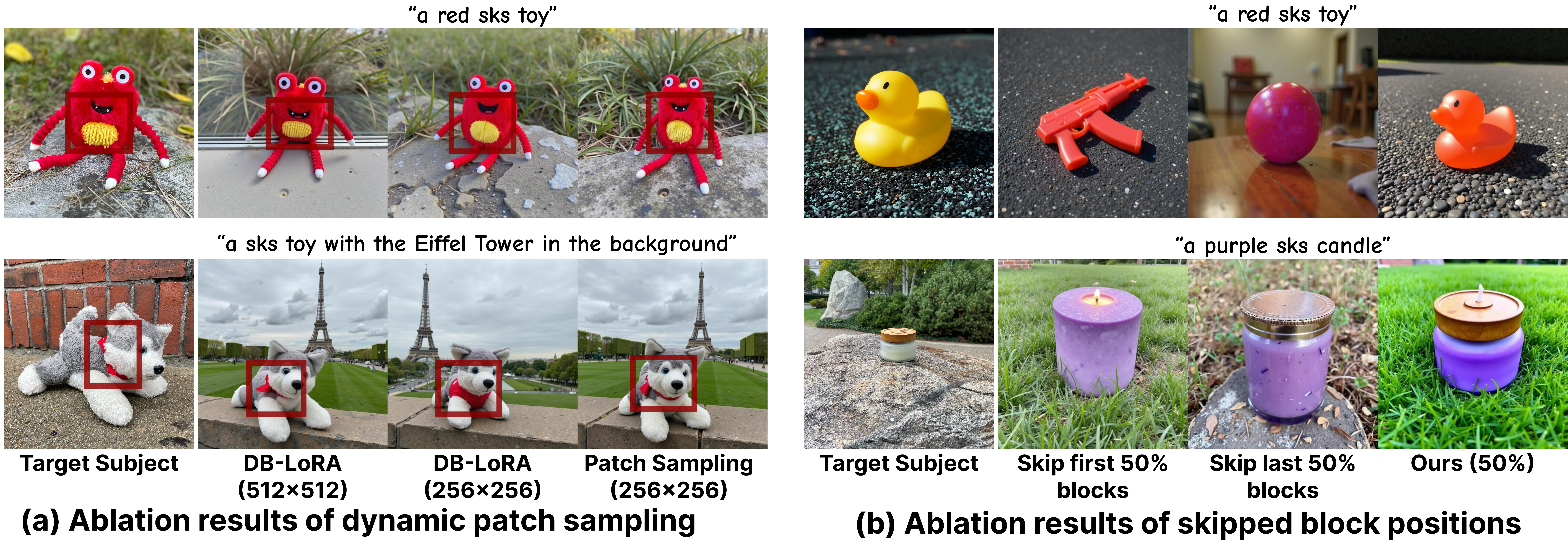}
    \vspace{-0.3cm}
    \caption{Qualitative ablation results of (a) dynamic patch sampling and (b) skipped block positions.}
    \vspace{-0.25cm}
    \label{fig:ablation_crop_and_skip_position}
\end{figure*}

\begin{figure*}[t!]
    \centering
    \includegraphics[width=0.92\linewidth]{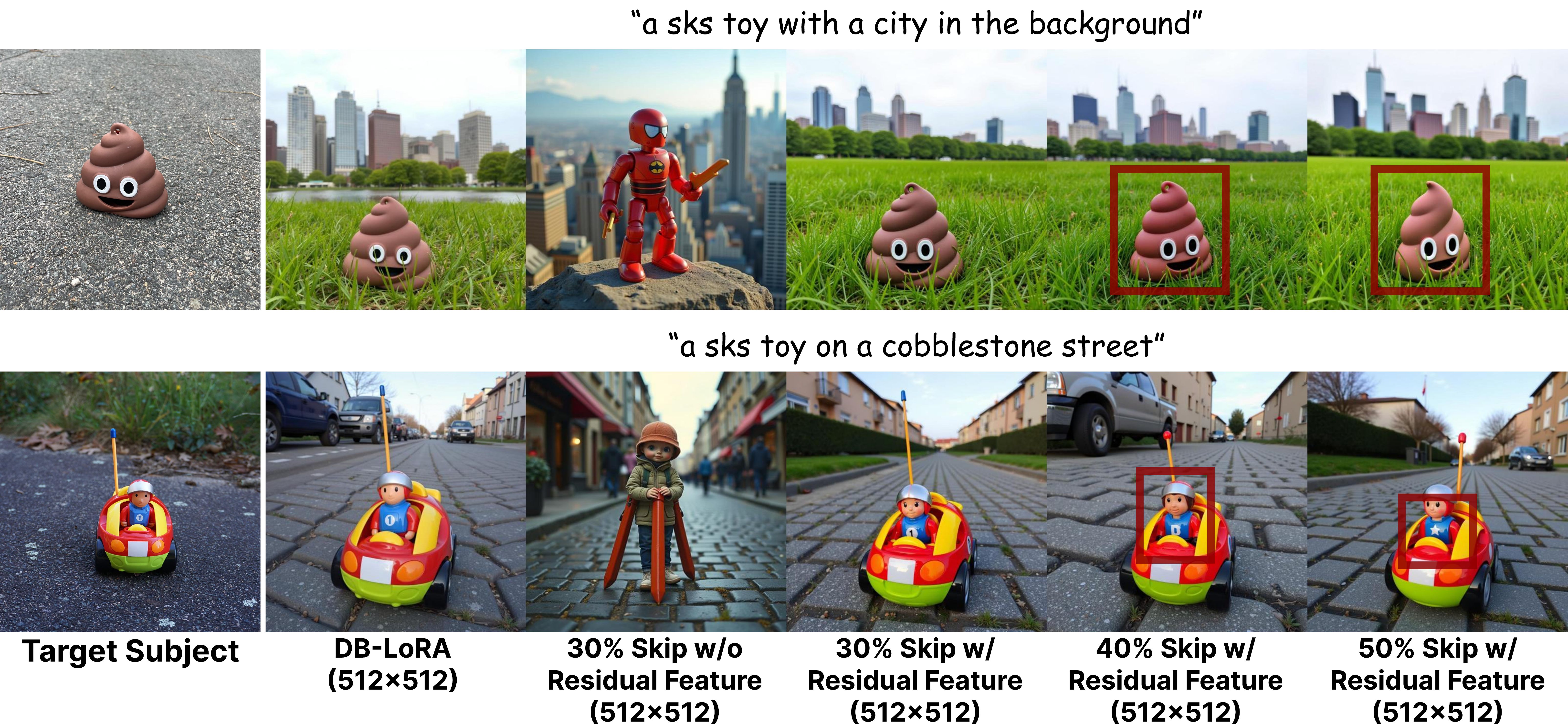}
    \vspace{-0.3cm}
    \caption{Qualitative ablation results of block skipping with and without residual features.}
    \vspace{-0.5cm}
    \label{fig:ablation_skip_ratio}
\end{figure*}

\noindent\textbf{User Study.}
Table~\ref{table:userstudy} presents the user preference results comparing LoRA and our method. 
In addition to reducing the training memory significantly, our approach achieves comparable subject and text fidelity. 
Further details are provided in the supplementary.

\noindent\textbf{Qualitative Results.}
Fig.~\ref{fig:qualitative} qualitatively compares the generation results of baselines and our method (30\% skip ratio).
Similar to the quantitative results, LoRA-FA and HollowedNet (30\% skip ratio) failed to generate the target subject. 
Notably, the comparison with the HollowedNet, which skips the middle blocks, demonstrates that the middle layers are crucial for personalization. 
Additionally, full fine-tuning DreamBooth shows decreased text adherence as seen in the second row of Fig.~\ref{fig:qualitative}. 
Our method, which uses approximately 58\% of the training memory compared to LoRA, demonstrates qualitatively comparable results.


\subsection{Ablation Study}
\noindent\textbf{Dynamic Patch Sampling.} Fig.~\ref{fig:ablation_crop_and_skip_position}(a) and Table~\ref{table:ablation_flux} compare our dynamic patch sampling with a standard fixed 256$\times$256 resizing. 
As shown, simply resizing often loses fine-grained details. 
Excessive downsampling can distort small regions (\textit{e.g.}, wolf plushie) or blur texture (\textit{e.g.}, monster toy), as highlighted by red boxes.
In contrast, our method effectively addresses these issues by leveraging crops sampled from lower timesteps, better preserving local detail. 
This leads to improved performance in DINO and CLIP-I.

\noindent\textbf{Effectiveness of Residual Feature and Block Skip.} 
Fig.~\ref{fig:ablation_skip_ratio} demonstrates the impact of residual feature precomputation.
Without residual features (third column), the model fails to personalize due to feature drift caused by skipping consecutive blocks.
Our residual precomputation alleviates this issue, maintaining personalization even under high skip ratios.
Compared to the upper bound (LoRA at 512 resolution; second column), skipping 30\%, 40\%, and 50\% of blocks still yields strong results.
Notably, the 30\% case performs nearly on par with LoRA fine-tuning, and even 50\% achieves comparable quality.
Table~\ref{table:ablation_flux} shows that our method maintains DINO and CLIP-I similar to LoRA fine-tuning while reducing training memory by 71\%.

\noindent\textbf{Ablation Study of Skipped Block Positions.} 
We evaluate the effect of applying residual precomputing at different block positions: first 50\%, last 50\%, and our proposed block selection under the same skip ratio of 50\% (Fig.\ref{fig:ablation_crop_and_skip_position}(b)).
The two baselines fail to preserve subject identity, highlighting the importance of mid-level blocks for capturing subject-specific features, as also discussed in Fig.~\ref{fig:skip_selection}.

\section{Conclusion}
This work addresses the crucial challenge of memory-efficient fine-tuning for DiT-based T2I models, which has become increasingly important for on-device personalization. 
We propose~\sysname, a memory-efficient fine-tuning framework that integrates dynamic patch sampling and block skipping.
Our method significantly reduces training memory consumption while maintaining the personalization performance. 
This work highlights the potential of combining spatial and architectural memory reduction strategies and opens the door for future exploration of memory-efficient training techniques tailored to transformer-based image generative models.

{
    \small
    \bibliographystyle{ieeenat_fullname}
    \bibliography{reference}
}

\clearpage
\setcounter{page}{1}
\maketitlesupplementary

\section{Implementation Details}

\subsection{Details of~\sysname}
\begin{table}[b!]
\centering
\caption{Skip indices selected in FLUX and SANA for the CustomConcept101 datasets. \textbf{Encoder} denotes the pre-trained encoder to extract image embedding for semantic distance.}
\resizebox{0.7\linewidth}{!}{
\begin{tabular}{c|c|c} 
\toprule
\textbf{Model-Encoder} & \textbf{Skip ratio}   & $(\mathbf{n^*, m^*})$    \\ 
\hline
\multirow{3}{*}{FLUX-DINO} & 30\%       &  (7, 10)   \\
& 40\%       & (13, 10)  \\
& 50\%       & (19, 10)  \\ 
\hline
\multirow{3}{*}{SANA-DINO} & 30\%       & (4, 2)    \\
& 40\%       & (4, 4)    \\
& 50\%       & (5, 5)    \\
\hline
\multirow{3}{*}{FLUX-CLIP} & 30\%       &  (7, 10)   \\
& 40\%       & (11, 12)  \\
& 50\%       & (18, 11)  \\ 
\hline
\multirow{3}{*}{SANA-CLIP} & 30\%       & (2, 4)    \\
& 40\%       & (4, 4)    \\
& 50\%       & (5, 5)    \\
                                  
\bottomrule
\end{tabular}
}
\vspace{0.3cm}
\label{supp-tab:skip_indices}
\end{table}
\noindent\textbf{Block Skipping with Residual Precomputing.}
The block skipping with residual precomputing is trained in two stages: precomputing and fine-tuning.
As it operates orthogonally to dynamic patch sampling, we apply dynamic patch sampling during the precomputing stage to extract and store residual features from images resized to a fixed lower resolution ($s_{\text{min}} \times s_{\text{min}}$). 
During fine-tuning, we apply the same patch sampling (\textit{e.g.}, location and timesteps) and diffusion timesteps during precomputing, to ensure correct alignment with the precomputed residual features.

To further validate the generalization capability of our skip indices selection algorithm, we selected skip indices for the DreamBooth experiments (Section~\ref{sec:experiment}) using the CustomConcept101 dataset. 
Specifically, we computed the indices of the skipped block for the randomly sampled 30 classes from CustomConcept101. 
To compute semantic distances between samples, we employed DINO~\cite{dino}. 
The values of n and m corresponding to the dataset and model configurations are summarized in Table~\ref{supp-tab:skip_indices}.
In our experiments, the FLUX and SANA models consist of 57 and 20 blocks, respectively. 
For FLUX, the number of skipped blocks $k$ was set to 17, 23, and 29 for skip ratios of 30\%, 40\%, and 50\%, respectively. 
For SANA, the corresponding values of $k$ were 6, 8, and 10.

\noindent\textbf{Hyperparameters.}
To ensure fair comparison with all baselines, our method was trained with a batch size of 1, a gradient accumulation step of 4, and 500 training iterations. 
The learning rate was set to 1e-4 with AdamW~\cite{adamw} optimizer, and the LoRA rank was set to 128 for FLUX and 512 for SANA, consistent with baseline settings. 
For inference, FLUX-based models used the Flow Match Euler Discrete scheduler~\cite{flux_scheduler} with a guidance scale of 3.5. 
SANA-based models followed the original implementation, employing Flow-DPM-Solver~\cite{dpmsolver} with 20 inference steps and a guidance scale of 4.5. 
Moreover, the models generated 1024$\times$1024 resolution images for evaluation.
All experiments were conducted using an NVIDIA A100 GPU.

\subsection{Details of Baselines}
All baseline implementations were based on the Diffusers library. 
For a fair comparison, all models were trained under the same settings: a batch size of 1, a gradient accumulation step of 4, and 500 training iterations using AdamW~\cite{adamw} optimizer.

\begin{table*}[t!]
\centering
\caption{Evaluation prompt list we used.}
\setlength{\tabcolsep}{5pt}
\small
\resizebox{1.0\linewidth}{!}{%
\begin{tabular}{c|c}
    \toprule
    \textbf{LIVE Prompt List} & \textbf{Non-LIVE Prompt List} \\
    \midrule
    \texttt{a <*> <subject> in the jungle} & \texttt{a <*> <subject> in the jungle} \\
    \texttt{a <*> <subject> in the snow} & \texttt{a <*> <subject> in the snow} \\
    \texttt{a <*> <subject> on the beach} & \texttt{a <*> <subject> on the beach} \\
    \texttt{a <*> <subject> on a cobblestone street} & \texttt{a <*> <subject> on a cobblestone street} \\
    \texttt{a <*> <subject> on top of pink fabric} & \texttt{a <*> <subject> on top of pink fabric} \\
    \texttt{a <*> <subject> on top of a wooden floor} & \texttt{a <*> <subject> on top of a wooden floor} \\
    \texttt{a <*> <subject> with a city in the background} & \texttt{a <*> <subject> with a city in the background} \\
    \texttt{a <*> <subject> with a mountain in the background} & \texttt{a <*> <subject> with a mountain in the background} \\
    \texttt{a <*> <subject> with a blue house in the background} & \texttt{a <*> <subject> with a blue house in the background} \\
    \texttt{a <*> <subject> on top of a purple rug in a forest} & \texttt{a <*> <subject> on top of a purple rug in a forest} \\
    \texttt{a <*> <subject> wearing a red hat} & \texttt{a <*> <subject> with a wheat field in the background} \\
    \texttt{a <*> <subject> wearing a santa hat} & \texttt{a <*> <subject> with a tree and autumn leaves in the background} \\
    \texttt{a <*> <subject> wearing a rainbow scarf} & \texttt{a <*> <subject> with the Eiffel Tower in the background} \\
    \texttt{a <*> <subject> wearing a black top hat and a monocle} & \texttt{a <*> <subject> floating on top of water} \\
    \texttt{a <*> <subject> in a chef outfit} & \texttt{a <*> <subject> floating in an ocean of milk} \\
    \texttt{a <*> <subject> in a firefighter outfit} & \texttt{a <*> <subject> on top of green grass with sunflowers around it} \\
    \texttt{a <*> <subject> in a police outfit} & \texttt{a <*> <subject> on top of a mirror} \\
    \texttt{a <*> <subject> wearing pink glasses} & \texttt{a <*> <subject> on top of the sidewalk in a crowded street} \\
    \texttt{a <*> <subject> wearing a yellow shirt} & \texttt{a <*> <subject> on top of a dirt road} \\
    \texttt{a <*> <subject> in a purple wizard outfit} & \texttt{a <*> <subject> on top of a white rug} \\
    \texttt{a red <*> <subject>} & \texttt{a red <*> <subject>} \\
    \texttt{a purple <*> <subject>} & \texttt{a purple <*> <subject>} \\
    \texttt{a shiny <*> <subject>} & \texttt{a shiny <*> <subject>} \\
    \texttt{a wet <*> <subject>} & \texttt{a wet <*> <subject>} \\
    \texttt{a cube shaped <*> <subject>} & \texttt{a cube shaped <*> <subject>} \\
    \bottomrule
\end{tabular}}
\label{table:prompt_list}
\end{table*}

\noindent\textbf{Textual Inversion~\cite{textual_inversion}.}
We used the official textual inversion implementation based on FLUX from the Diffusers library.
We set the number of trainable tokens to four and the learning rate to 1e-3.
Although various hyperparameter settings were explored, we report the results using the configuration that yielded the most reasonable performance. 
Nevertheless, we observed that textual inversion alone was insufficient for achieving satisfactory personalization with DiT-based T2I models.
Since SANA does not provide a dedicated implementation for textual inversion, we implemented it manually by referencing the FLUX-based codebase. 
Specifically, in the PyTorch implementation, the embedding layer was included in the optimizer, and at each training iteration, all tokens except the trainable ones were overwritten with their original values. 
Although including only the trainable tokens in the optimizer could reduce the optimizer state memory footprint, we adhered to the official FLUX-based textual inversion implementation provided by the Diffusers library for consistency in our experiments. 
Consequently, as shown in Table~\ref{supp-table:SANA-memory}, SANA, which employs a large language model as its text encoder, requires a substantial optimizer state memory due to the embedding layer containing approximately 260,000 tokens.
Nevertheless, a significant drawback of this approach is that the text encoder must remain loaded on the GPU throughout training, and the backpropagation process must extend all the way to the embedding layer of the text encoder. 
As a result, this method incurs substantially higher memory consumption compared to approaches such as LoRA or our proposed method.

\noindent\textbf{DreamBooth~\cite{dreambooth}.}
We utilized the official DreamBooth implementation from the Diffusers library. 
A learning rate of 5e-5 was used for fine-tuning FLUX, and 1e-4 for SANA. 
Prior preservation loss was also employed during training.

\noindent\textbf{LoRA~\cite{lora}.}
We used the official LoRA implementation from the Diffusers library. 
We used a learning rate of 1e-4 and set the LoRA rank to 128 for FLUX and 512 for SANA. 

\noindent\textbf{LISA~\cite{lisa}.}  
We implemented LISA by referencing the original paper and modifying the Diffusers LoRA codebase accordingly. 
All experiments were conducted using the same LoRA configuration and hyperparameters for fair comparison.

\noindent\textbf{LoRA-FA~\cite{lorafa}.}  
LoRA-FA was implemented based on its original paper using the Diffusers LoRA code as a foundation. We ensured consistent LoRA settings and hyperparameters to maintain a fair experimental setup.

\noindent\textbf{HollowedNet~\cite{hollowednet}.}
As the official implementation was not publicly available, we reimplemented HollowedNet based on the paper and guidance from the authors. 
The original method was demonstrated only on U-Net-based Stable Diffusion~\cite{ldm}; thus, we modified it to support transformer block skipping by incorporating our residual feature precomputing method. 
Since HollowedNet fine-tunes only shallow layers in U-Net, we applied a similar strategy to DiT-based T2I models. 
For the fair comparison, we used the same hyperparameter settings including the learning rate, batch size, and rank as those used for LoRA.

\subsection{Prompt List for Evaluation.}
Table~\ref{table:prompt_list} shows the prompt list we utilized to generate images for evaluation.
The text prompts are identical to the text prompts provided in the DreamBooth dataset~\cite{dreambooth}.

\subsection{Details of User Study}
We assessed user preferences between LoRA~\cite{lora}, which shows the best performance across the baselines, and our proposed model based on FLUX.
Participants were instructed to select the image with better subject and text fidelity, or choose ‘undecided’ if the two results were comparable.
Specifically, subject fidelity was evaluated based on similarity to the training image, while text fidelity was judged by how well the image reflected the input prompt.
The user study was conducted over 60 sets.
The final results were calculated by averaging the preferences across all participants.

\begin{table*}[t!]
\centering
\caption{Detailed training memory comparison and TFLOPs with baselines based on \textbf{FLUX}. Inference resolution is $1024\times1024$}
\setlength{\tabcolsep}{3.5pt} 
\small
\begin{tabular}{l|cc|c|ccc|cc}
    \toprule
    \multirowcell{2}{\textbf{Method}} & \multirowcell{2}{\textbf{Skip} \\ \textbf{Ratio}} & \multirowcell{2}{\textbf{Training} \\ \textbf{Resolution}} & \multirowcell{2}{\textbf{Training} \\ \textbf{Memory}} & \multirowcell{2}{\textbf{Parameter} \\ \textbf{Memory}} & \multirowcell{2}{\textbf{Optimizer} \\ \textbf{State Memory}} & \multirowcell{2}{\textbf{Other} \\ \textbf{Memory}} & \multirowcell{2}{\textbf{
    Training} \\ \textbf{TFLOPs}} & \multirowcell{2}{\textbf{
    Precompute} \\ \textbf{TFLOPs}} \\
    & & & & & & \\
    \midrule
    Textual Inversion   & -- & 512$\times$512 & 42.15 GiB & 31.27 GiB & \textbf{0.14} GiB & 10.74 GiB & 44.57 & -- \\
    DreamBooth & -- & 512$\times$512 & 68.96 GiB & 22.17 GiB & 22.52 GiB & 24.27 GiB & 158.77 & --\\    
    LoRA       & -- & 512$\times$512 & 35.99 GiB & 22.84 GiB & 1.34 GiB & 11.82 GiB & 41.67 & -- \\
    LISA       & -- & 512$\times$512 & 35.98 GiB & 22.84 GiB & 1.34 GiB & 9.33 GiB & 41.67 & -- \\
    LoRA-FA    & -- & 512$\times$512 & 32.48 GiB & 22.84 GiB & 0.67 GiB & 8.97 GiB & 41.32 & -- \\
    \midrule
    HollowedNet & 30 \% & 512$\times$512 & 24.11 GiB & 15.62 GiB & 1.04 GiB & 7.46 GiB & 24.35 & 19.85 \\
    HollowedNet & 40 \% & 512$\times$512 & 18.40 GiB & 11.91 GiB & 0.79 GiB & 5.70 GiB & 18.44  & 19.85 \\
    HollowedNet & 50 \% & 512$\times$512 & \underline{12.23} GiB & \underline{7.93} GiB & 0.54 GiB & 3.76 GiB & \underline{11.80} & 19.85 \\    
    \midrule
    Ours   & 30 \% & 256$\times$256 & 20.78 GiB & 15.54 GiB & 0.88 GiB & 4.37 GiB & 14.59 & 9.94 \\
    Ours   & 40 \% & 256$\times$256 & 15.61 GiB & 11.58 GiB & 0.56 GiB & \underline{3.46} GiB & 12.30 & 9.94 \\
    Ours   & 50 \% & 256$\times$256 & \textbf{10.42} GiB & \textbf{7.63} GiB & \underline{0.25} GiB & \textbf{2.55} GiB & \textbf{9.96} & 9.94 \\
    \bottomrule
\end{tabular}
\label{supp-table:FLUX-memory}
\end{table*}

\begin{table*}[t!]
\centering
\caption{Detailed training memory comparison and TFLOPs with baselines based on \textbf{SANA}. Inference resolution is $1024\times1024$}
\setlength{\tabcolsep}{3.5pt} 
\small
\begin{tabular}{l|cc|c|ccc|cc}
    \toprule
    \multirowcell{2}{\textbf{Method}} & \multirowcell{2}{\textbf{Skip} \\ \textbf{Ratio}} & \multirowcell{2}{\textbf{Training} \\ \textbf{Resolution}} & \multirowcell{2}{\textbf{Training} \\ \textbf{Memory}} & \multirowcell{2}{\textbf{Parameter} \\ \textbf{Memory}} & \multirowcell{2}{\textbf{Optimizer} \\ \textbf{State Memory}} & \multirowcell{2}{\textbf{Other} \\ \textbf{Memory}} & \multirowcell{2}{\textbf{
    Training} \\ \textbf{TFLOPs}} & \multirowcell{2}{\textbf{
    Precompute} \\ \textbf{TFLOPs}} \\
    & & & & & & \\
    \midrule
    Textual Inversion   & -- & 1024$\times$1024 & 15.31 GiB & 7.86 GiB & 2.20 GiB & 5.25 GiB & 8.91 & -- \\
    DreamBooth          & -- & 1024$\times$1024 & 24.69 GiB & 2.99 GiB & 5.98 GiB & 15.72 GiB & 13.34 & -- \\    
    LoRA                & -- & 1024$\times$1024 & 8.35 GiB & 3.50 GiB & 1.03 GiB & 3.83 GiB & 6.41 & -- \\
    LISA                & -- & 1024$\times$1024 & 6.67 GiB & 3.50 GiB & \textbf{0.10} GiB & 3.07 GiB & 6.00 & -- \\
    LoRA-FA             & -- & 1024$\times$1024 & 7.40 GiB & 3.50 GiB & \underline{0.51} GiB & 3.39 GiB & 6.16 & -- \\
    \midrule
    HollowedNet     & 30 \% & 1024$\times$1024 & 5.90 GiB & 2.48 GiB & 0.72 GiB & 2.70 GiB & 4.48 & 2.91 \\    
    HollowedNet     & 40 \% & 1024$\times$1024 & 5.08 GiB & \underline{2.14} GiB & 0.62 GiB & 2.33 GiB & 3.84 & 2.91 \\    
    HollowedNet     & 50 \% & 1024$\times$1024 & 4.26 GiB & \textbf{1.79} GiB & \underline{0.51} GiB & 1.95 GiB & 3.19 & 2.91 \\    
    \midrule
    Ours       & 30 \% & 512$\times$512 & 4.29 GiB & 2.48 GiB & 0.72 GiB & 1.09 GiB & 1.26 & 0.82 \\
    Ours       & 40 \% & 512$\times$512 & \underline{3.69} GiB & \underline{2.14} GiB & 0.62 GiB & \underline{0.94} GiB & \underline{1.08} & 0.82 \\
    Ours       & 50 \% & 512$\times$512 & \textbf{3.10} GiB & \textbf{1.79} GiB & \underline{0.51} GiB & \textbf{0.79} GiB & \textbf{0.90} & 0.82 \\
    \bottomrule
\end{tabular}
\label{supp-table:SANA-memory}
\end{table*}

\section{Training Memory}\label{sec:supple_training_memory}

Table~\ref{supp-table:FLUX-memory} and~\ref{supp-table:SANA-memory} report detailed training memory usage for FLUX- and SANA-based baselines as well as our proposed method. 
To isolate memory usage relevant to trainable components, we precomputed text features when the text encoder was not updated and encoded training images into latent space using a VAE in advance; both were excluded from memory calculations (except for textual inversion). 
We conducted training using \texttt{bfloat16} mixed precision, which is commonly adopted for FLUX and SANA for memory-efficient training.

As shown in the tables, we categorize training memory into three components: (1) \textbf{Parameter Memory}, (2) \textbf{Optimizer State Memory}, and (3) \textbf{Other Memory}, which includes the combined memory from forward (activations) and backward (gradients), excluding parameters and optimizer states. 
Temporary memory was negligible and thus not considered in our measurements.

As shown in Tables~\ref{supp-table:FLUX-memory} and~\ref{supp-table:SANA-memory}, the proposed method, demonstrates the lowest consumption of training memory and TFLOPs among the compared approaches. 
In the case of FLUX, the presence of two types of transformer blocks, such as single attention block and dual attention block, results in differing block indices for skipping, which leads to discrepancies in parameter memory usage compared to other HollowedNet variants. 
Notably, the technique of block skipping with residual feature precomputing significantly reduces parameter memory, optimizer state memory, and other memory. 
Furthermore, dynamic patch sampling, while not affecting parameter or optimizer state memory, substantially decreases both forward and backward memory usage as well as TFLOPs.

In contrast, textual inversion requires continuous forward and backward passes through the text encoder during training, preventing precomputation of text prompts (\textit{e.g.}, "a photo of sks dog") and thereby increasing parameter memory requirements. 
PEFT methods such as LISA and LoRA-FA reduce optimizer state memory but do not significantly impact overall training memory. 
Although HollowedNet achieves considerable memory savings, it suffers from a substantial performance drop, as described in the main paper. 
As it is based on a U-Net architecture, we integrated our residual feature precomputing technique with HollowedNet. 
However, due to the empirical nature of U-Net-based methods, HollowedNet fails to identify crucial layers for personalization.

\begin{table*}[t!]
\centering
\small
\caption{Comparison with baselines based on \textbf{FLUX} in CustomConcept101 dataset. Inference resolution is $1024\times1024$}
\begin{tabular}{l|cc|c|ccc}
    \toprule
    \multicolumn{1}{c|}{\textbf{Method}} & \textbf{Skip Ratio} & \textbf{Training Res.} & \textbf{Training Mem.} & \textbf{DINO} & \textbf{CLIP-I} & \textbf{CLIP-T} \\
    \midrule
    LoRA                & - & 512$\times$512 & 35.99 GiB & \textbf{0.6726} & \textbf{0.7961} & 0.2937 \\
    LISA                & - & 512$\times$512 & 35.98 GiB & 0.5985 & 0.7491 & \textbf{0.3085} \\
    \midrule
    HollowedNet     & 30 \% & 512$\times$512 & 24.11 GiB & 0.4378 & 0.6570 & 0.3021 \\    
    HollowedNet     & 40 \% & 512$\times$512 & 18.40 GiB & 0.4568 & 0.6662 & \underline{0.3073} \\    
    HollowedNet     & 50 \% & 512$\times$512 & \underline{12.23} GiB & 0.4242 & 0.6553 & 0.3061 \\
    \midrule
    Ours                & 30 \% & 256$\times$256 & 20.78 GiB & \underline{0.6455} & \underline{0.7737} & 0.3004 \\
    Ours                & 40 \% & 256$\times$256 & 15.61 GiB & 0.6377 & 0.7696 & 0.3007 \\
    Ours                & 50 \% & 256$\times$256 & \textbf{10.42} GiB & 0.6137 & 0.7513 & 0.3056 \\
    \bottomrule
\end{tabular}
\label{supp-table:custom101_flux}
\end{table*}

\begin{table*}[t!]
\centering
\caption{Comparison with baselines based on \textbf{SANA} in CustomConcept101 dataset. Inference resolution is $1024\times1024$}
\small
\begin{tabular}{l|cc|c|ccc}
    \toprule
    \multicolumn{1}{c|}{\textbf{Method}} & \textbf{Skip Ratio} & \textbf{Training Res.} & \textbf{Training Mem.} & \textbf{DINO} & \textbf{CLIP-I} & \textbf{CLIP-T} \\
    \midrule
    LoRA                 & - & 1024$\times$1024 & 8.35 GiB & \textbf{0.6545} & 0.7792 & \underline{0.3096} \\
    LISA                & - & 1024$\times$1024 & 6.67 GiB & 0.4944 & 0.6970 & \textbf{0.3213} \\
    \midrule
    HollowedNet   & 30 \% & 1024$\times$1024 & 5.90 GiB & 0.6459 & 0.7708 & 0.2983 \\
    HollowedNet   & 40 \% & 1024$\times$1024 & 5.08 GiB & 0.6257 & 0.7537 & 0.2932 \\
    HollowedNet   & 50 \% & 1024$\times$1024 & 4.26  GiB & 0.5828 & 0.7191 & 0.2955 \\
    \midrule
    Ours                & 30 \% & 512$\times$512 & 4.29 GiB & \underline{0.6522} & \textbf{0.7826} & 0.3040 \\
    Ours                & 40 \% & 512$\times$512 & \underline{3.69} GiB & 0.6489 & \underline{0.7810} & 0.3025 \\
    Ours                & 50 \% & 512$\times$512 & \textbf{3.10} GiB & 0.6441 & 0.7765 & 0.3033 \\
    \bottomrule
\end{tabular}
\label{supp-table:custom101_sana}
\end{table*}

\section{Additional Results} 

\subsection{CustomConcept101 Results}\label{supp-sec:custom101}
Table~\ref{supp-table:custom101_flux} and Table~\ref{supp-table:custom101_sana} present the quantitative results on the CustomConcept101 dataset. 
We include only the models that performed competitively in the DreamBooth setting, excluding Textual Inversion, DreamBooth, and LoRA-FA due to their relatively low performance. 
Similar to the results on the DreamBooth dataset, our method achieves performance comparable to LoRA.
Notably, SANA (CLIP-I) achieves the highest performance even while skipping 30\% of transformer blocks. In contrast, LISA and HollowedNet exhibit a significant performance drop when 50\% of the blocks are skipped.

\begin{figure*}[h!]
    \centering
    \includegraphics[width=1.0\linewidth]{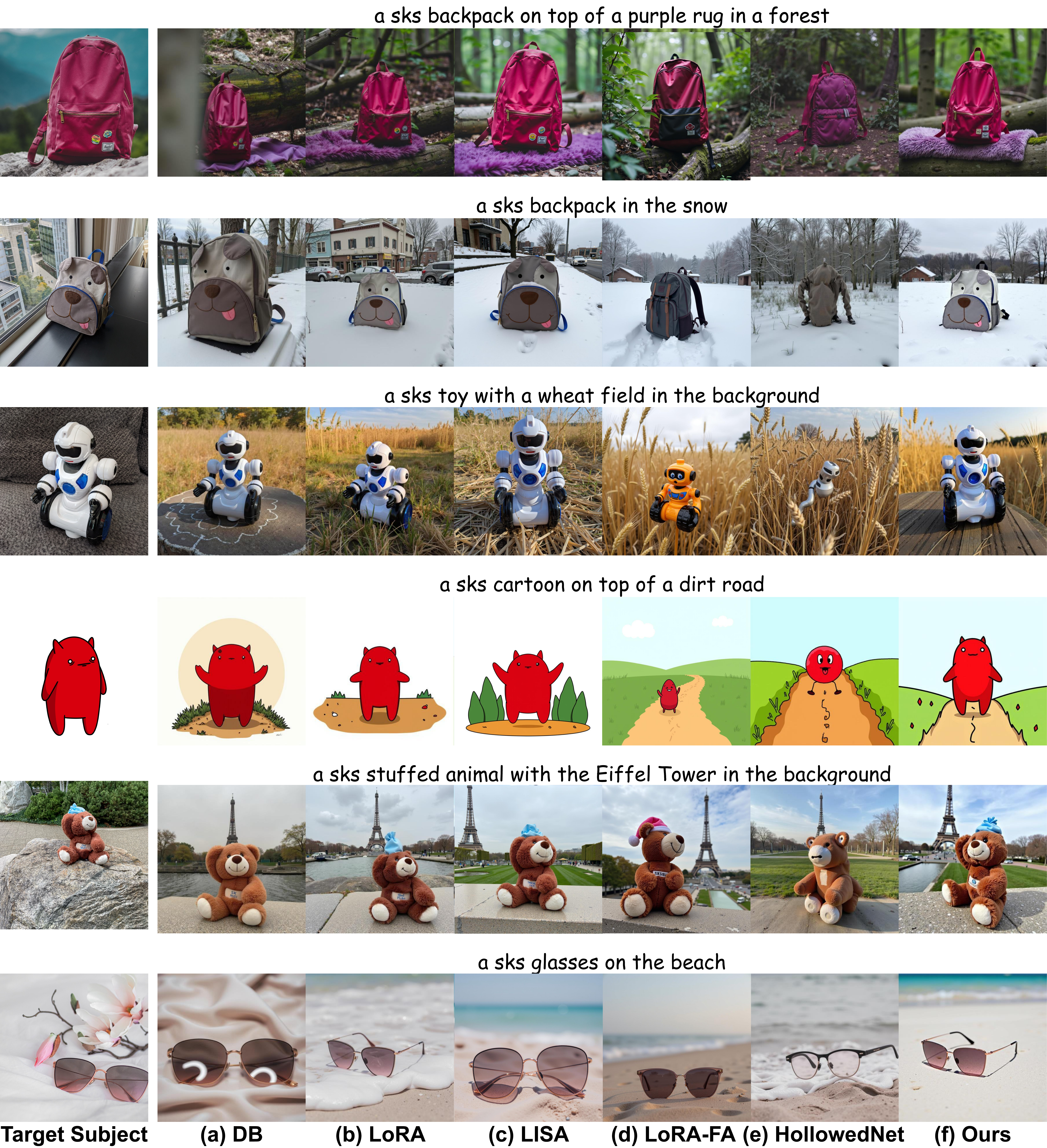}
    \caption{Additional qualitative comparison with baselines on the DreamBooth dataset.}
    \label{supp-fig:qualitative_db}
\end{figure*}

\begin{figure*}[h!]
    \centering
    \includegraphics[width=0.8\linewidth]{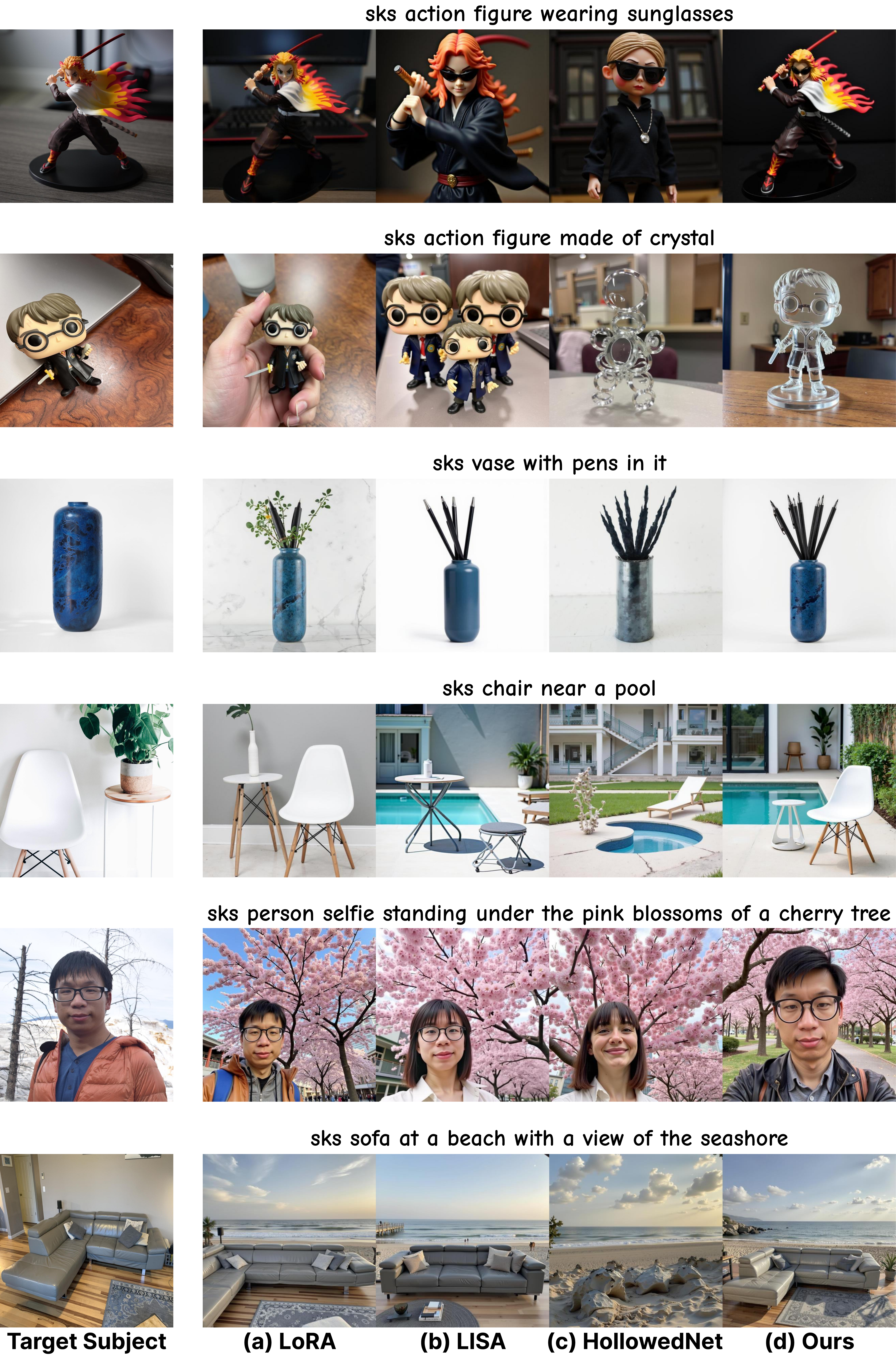}
    \caption{Qualitative comparison with baselines on the CustomConcept101 dataset.}
    \label{supp-fig:qualitative_cutom101}
\end{figure*}

Fig.~\ref{supp-fig:qualitative_db} and Fig.~\ref{supp-fig:qualitative_cutom101} provide qualitative comparisons between our method and FLUX-based baseline models on the DreamBooth and CustomConcept101 datasets, respectively. 
In particular, as shown in Fig.~\ref{supp-fig:qualitative_cutom101}, while LISA fails to preserve the identity of the target subject, our method demonstrates both superior memory efficiency and competitive identity preservation and text fidelity compared to LoRA.

\begin{table}[t!]
\centering
\caption{Comparison with partial LoRA and gradient checkpointing using SANA.}
\label{supp-tab:comparison_with_partial_lora}
\small
\addtolength{\tabcolsep}{-3.5pt}
\begin{tabular}{@{}lcccc@{}}
\toprule
Method & Training Mem. & DINO$^\uparrow$ & CLIP-I$^\uparrow$ & CLIP-T$^\uparrow$ \\ 
\midrule
LoRA & 8.35 GiB & 0.7374 & 0.8108 & 0.3254 \\
Grad. Checkpointing & 5.66 GiB & 0.7374 & 0.8122 & 0.3241 \\
Partial LoRA (50\%) & 7.20 GiB & 0.6870 & 0.7874 & 0.3261 \\
\textbf{Ours (50\%)} & \textbf{3.10 GiB} & 0.7277 & 0.8034 & 0.3177 \\
\bottomrule
\end{tabular}
\end{table}

\begin{figure*}[t!]
    \centering
    \includegraphics[width=1.0\linewidth]{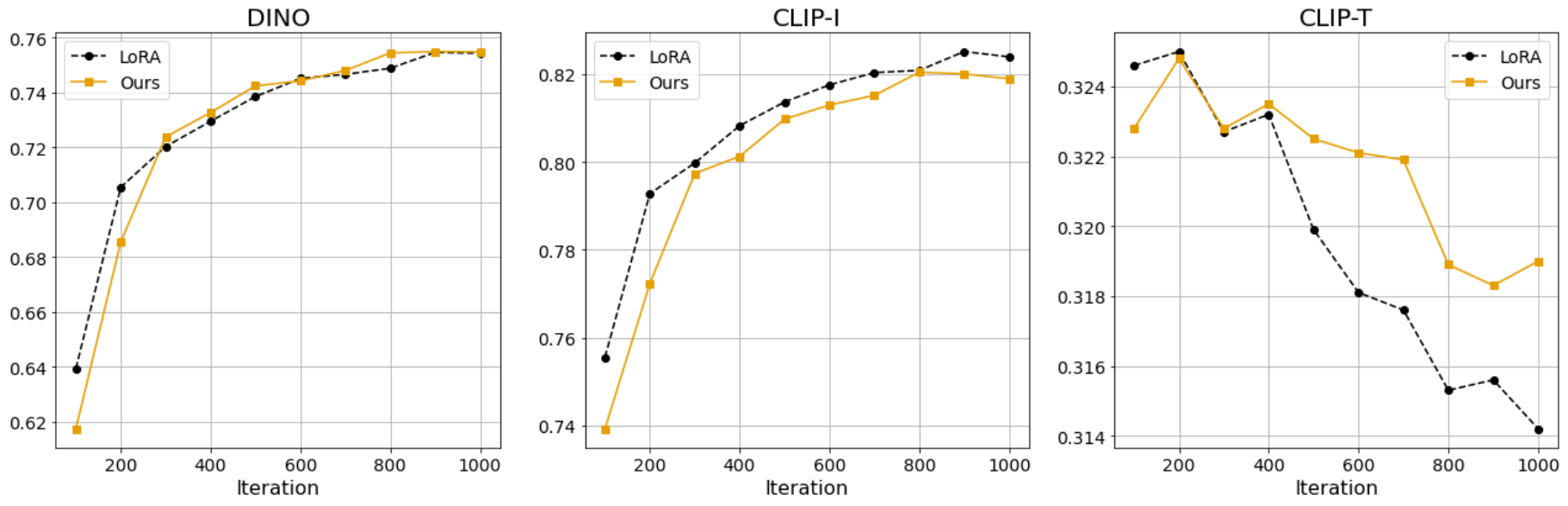}
    \caption{Performance comparison across training iterations for LoRA and our method.}
    \label{supp-fig:model_convergence}
\end{figure*}

\subsection{Comparison with Additional Baselines}

We further evaluate our approach against two widely used memory-efficient baselines: gradient checkpointing and partial LoRA. 
Gradient checkpointing reduces peak activation memory by discarding intermediate activations during the forward pass and recomputing them only when needed during the backward pass. 
Partial LoRA mitigates the memory overhead of optimizer states by selectively fine-tuning only a subset of layers. 
For partial LoRA (50\%), we selectively fine-tune only the 50\% of blocks selected by our block selection algorithm.
Table~\ref{supp-tab:comparison_with_partial_lora} shows the results of partial LoRA and gradient checkpointing using SANA model.
While these techniques are straightforward to implement for memory reduction, they both necessitate that the entire base model parameters remain resident in GPU VRAM. 
In contrast, our method utilizes block skipping to actively offload inactive weights from the GPU, significantly reducing the weight memory footprint of SANA while maintaining comparable performance.

\subsection{Ablation Results on Dynamic Patch Sampling}

Our dynamic patch sampling mechanism adopts a `Low-to-High' strategy, where the patch size is increased in proportion to the diffusion timestep $t$ (i.e., patch size $\uparrow$ as $t \uparrow$), utilizing larger patches at higher noise levels to align with the structural-to-textural transition of the generative process. 
This principled approach captures coarse global structures via larger patches and refines fine details via smaller patches. 
To validate its effectiveness, we conducted additional ablation study using the FLUX model, comparing our method against `High-to-Low' (patch size $\downarrow$ as $t \uparrow$) and `Random' sampling baselines. 
As shown in Table~\ref{supp-tab:ablation_patch_sampling}, our strategy consistently achieves superior performance across all evaluation metrics, highlighting the advantage of our proposed dynamic patch sampling.

\begin{table}[h!]
\centering
\caption{Comparison of different patch sampling using FLUX.}
\resizebox{1\linewidth}{!}{
\addtolength{\tabcolsep}{-3.5pt}
\begin{tabular}{@{}lcccc@{}}
\toprule
\textbf{Method (FLUX)} & \textbf{Training Res.} & \textbf{DINO$^\uparrow$} & \textbf{CLIP-I$^\uparrow$} & \textbf{CLIP-T$^\uparrow$} \\ 
\midrule
High-to-Low & $256\times256$ & 0.7164 & 0.8093 & 0.3182 \\
Random & $256\times256$ & 0.7187 & 0.8068 & 0.3192 \\ 
\textbf{Ours} & $256\times256$ & \textbf{0.7253} & \textbf{0.8099} & \textbf{0.3196} \\
\bottomrule
\end{tabular}}
\label{supp-tab:ablation_patch_sampling}
\end{table}

\subsection{Ablation Results on Memory Consumption}
We provide a detailed ablation study on the memory consumption of our method, comparing the effects of Dynamic Patch Sampling and Block Skipping. 
As shown in Table~\ref{supp-tab:additional_memory_ablation}, the baseline LoRA model requires 35.99 GiB of memory. 
Applying only Dynamic Patch Sampling reduces the memory usage to 30.48 GiB. 
In contrast, applying Block Skipping alone at levels of 30\%, 40\%, and 50\% progressively reduces memory usage to 24.24 GiB, 18.84 GiB, and 13.40 GiB, respectively. 
Both methods individually reduce memory usage while maintaining performance. 
When used together, they offer further improvements in training memory efficiency.

\begin{table}[h!]
\centering
\caption{Memory usage with dynamic patch sampling and block skipping applied.}
\resizebox{1\linewidth}{!}{
\small
\addtolength{\tabcolsep}{-3.5pt}
\begin{tabular}{lcc} 
\toprule
\textbf{Method (FLUX)} & \textbf{Training Res.} & \textbf{Memory (GiB)}  \\ 
\hline
LoRA                                                 & 512x512             & 35.99               \\
\textcolor[rgb]{0.2,0.2,0.2}{Dynamic Patch Sampling} & 256x256             & 30.48               \\
\textcolor[rgb]{0.2,0.2,0.2}{Block Skipping (30\%)}  & 512x512             & 24.63               \\
\textcolor[rgb]{0.2,0.2,0.2}{Block Skipping (40\%)}  & 512x512             & 18.76               \\
\textcolor[rgb]{0.2,0.2,0.2}{Block Skipping (50\%)}  & 512x512             & 12.85               \\
\textcolor[rgb]{0.2,0.2,0.2}{Ours (30\%)}          & 256x256             & 20.78               \\
\textcolor[rgb]{0.2,0.2,0.2}{Ours (40\%)}          & 256x256             & 15.61               \\
\textcolor[rgb]{0.2,0.2,0.2}{Ours (50\%)}          & 256x256             & 10.42               \\
\bottomrule
\end{tabular}}
\label{supp-tab:additional_memory_ablation}
\end{table}

\subsection{Training Dynamics and Convergence Behavior}
All methods were trained for the same number of iterations (500 iterations as reported in the main paper). 
To assess convergence behavior in detail, we additionally trained SANA using both LoRA and our method for 1,000 iterations, and report performance measured every 100 iterations in Fig.~\ref{supp-fig:model_convergence}.

While LoRA exhibits slightly better performance in the early stages (up to 200 iterations), it fails to fully converge. 
Beyond 300 iterations, both methods continue to improve in terms of image fidelity, as reflected in the DINO and CLIP-I scores, and ultimately achieve comparable performance. 
These results demonstrate that our method converges at a similar rate to LoRA while maintaining competitive performance throughout the training process.

\begin{figure*}[t!]
    \centering
    \includegraphics[width=1.0\linewidth]{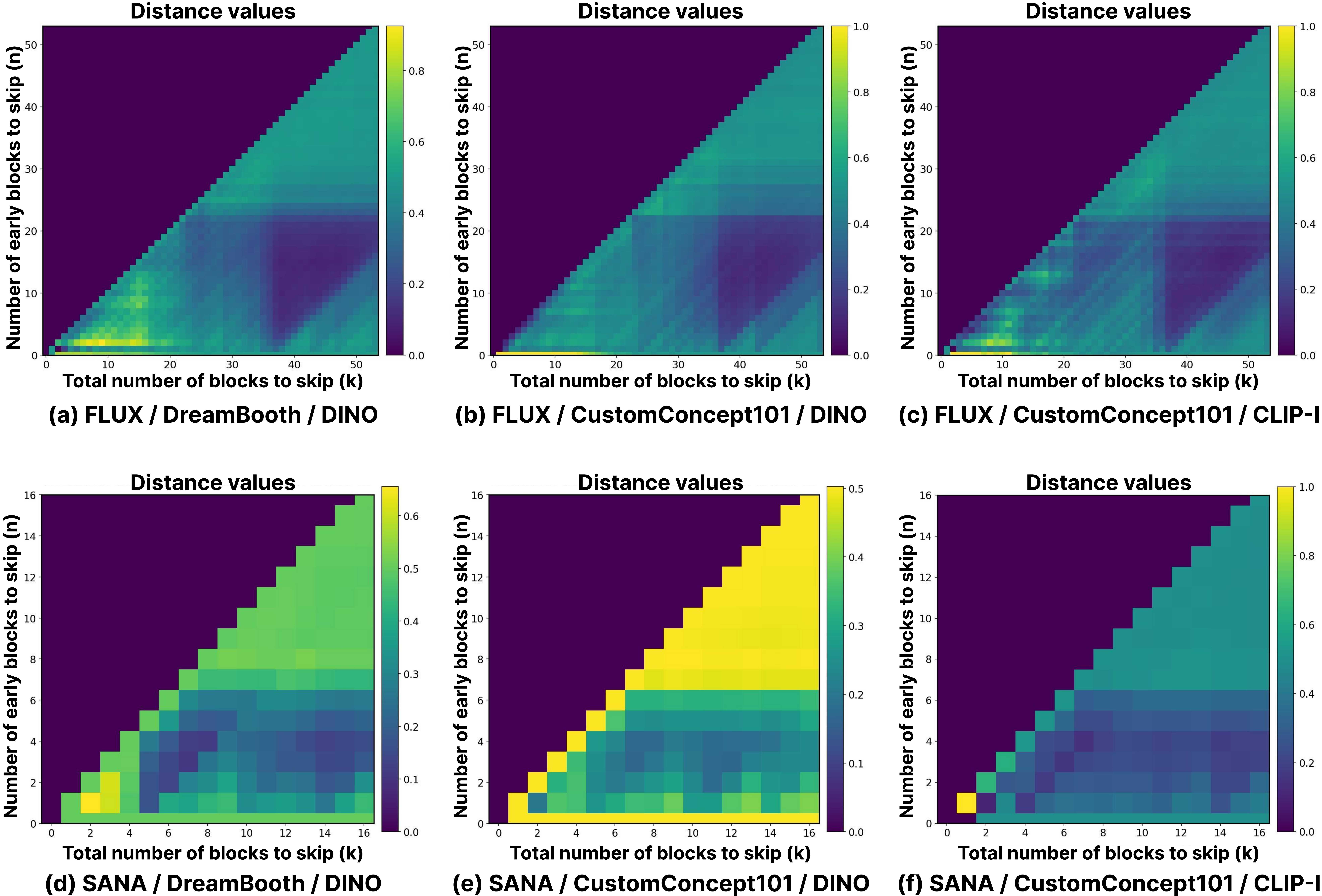}
    \vspace{-0.4cm}
    \caption{Visualization of average semantic distances for all combinations of $n$ and $m$. Each subplot is labeled with the corresponding model, dataset, and pre-trained encoder. Since $m=k-n$, each column shares the same total number of skipped blocks $k$. Each value shows the average distance when cross-attention is masked on the first $n$ and the last $m$ blocks.}
    \vspace{-0.4cm}
    \label{supp-fig:dist_value}
\end{figure*}

\begin{figure}[h!]
    \centering
    \includegraphics[width=1.0\linewidth]{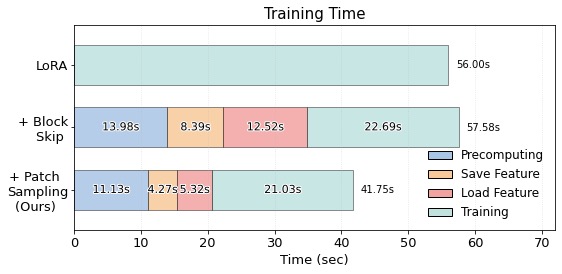}
    \caption{Comparison of training time between LoRA and our method including block skipping and dynamic patch sampling.}
    \label{supp-fig:training-time}
\end{figure}

\subsection{Training Time}
We analyze the training time of our method with a 50\% block skipping ratio applied to the SANA model over 100 iterations.
Fig.~\ref{supp-fig:training-time} depicts the training time of our method relative to LoRA, highlighting the impact of block skipping and dynamic patch sampling.
As shown in Fig.~\ref{supp-fig:training-time}, we break down the total time spent across four stages: precomputing, saving features, loading features, and training. 
Even when accounting for precomputation and I/O overhead, our final method (41.75 s) achieves approximately 25\% faster training time compared to LoRA (56.00 s). 
Furthermore, for FLUX, our method reduces training memory usage by about 71\% compared to the baseline, which is the primary contribution of this work.

\subsection{Semantic Distances}
Figure~\ref{supp-fig:dist_value} visualizes the semantic distance for all combinations of $n$ and $m$ on the DreamBooth and CustomConcept101 datasets using the FLUX and SANA models. 
Each value represents the average semantic distance when the image-to-text cross-attention maps in the first $n$ and last $m$ blocks are masked.
Fig.~\ref{supp-fig:dist_value} shows that semantic distance increases when $n$ exceeds approximately 23 in FLUX and 8 in SANA. 
This suggests that the middle blocks, when skipped, contribute more to the degradation of output quality, underscoring their importance in preserving subject identity.
Furthermore, we conducted an ablation study on the encoder used for computing the proposed semantic distance. 
Table~\ref{supp-tab:skip_indices} shows that the index differences are minimal when using DINO similarity (DINO) versus CLIP image similarity (CLIP-I), showing that both measures yield similar semantic distance trends. 
CLIP text similarity (CLIP-T) was excluded from the analysis because it lacks the ability to accurately capture subject identity.
This observation is further supported by Fig.~\ref{supp-fig:dist_value} (b) and (c), as well as (e) and (f), where the patterns of semantic distance are consistent across DINO and CLIP-I scores.

\section{Discussion and Limitations}
\noindent\textbf{Reducing Memory for Residual Feature precomputing.}
For both our method and HollowedNet~\cite{hollowednet}, we excluded the memory used during the precomputing stage from the reported training memory. 
This is because, in DiT-based models, it is feasible to partially forward only specific layers. 
In particular, FLUX alone requires approximately 22.17~GiB of parameter memory, making it impractical to load the entire model onto the GPU simultaneously in on-device personalization scenarios.

To address this, we can perform partial forward passes for residual feature precomputing: only the necessary transformer blocks are loaded into the GPU to compute and store intermediate feature maps up to that point.
The remaining blocks are then loaded sequentially to complete the precomputing of residual features. 
This staged strategy significantly reduces parameter memory usage during forward computation, which otherwise constitutes the largest portion of memory consumption.

\noindent\textbf{Compatibility with Other Methods.}  
To reduce training memory, prior works have leveraged techniques such as quantization, parameter-efficient fine-tuning (PEFT) methods like LoRA~\cite{lora}, and gradient checkpointing. 
Our method is compatible with these approaches and can be integrated to further reduce memory overhead. 
For example, as demonstrated by our integration with LoRA, our method can be extended to other PEFT techniques, such as DoRA~\cite{dora}, without loss of generality.

\noindent\textbf{On-Device Deployment.}
Although actual demonstration on mobile or IoT device falls outside our current scope, which focuses on the fundamental algorithmic efficiency of DiT fine-tuning, our substantial reductions in VRAM and FLOPs significantly enhance edge feasibility. 
A practical consideration for on-device deployment is the non-trivial ROM required for storing precomputed features. 
While this overhead can be mitigated by adopting a periodic feature-saving strategy (\textit{e.g.}, caching per iteration) rather than full-dataset pre-storage, such an approach introduces additional implementation complexity. 
We consider the co-optimization of ROM and VRAM for seamless on-device fine-tuning as a promising direction for future work.

\end{document}